\documentclass{article}

\PassOptionsToPackage{square, numbers}{natbib}

\usepackage[final]{neurips_2022}

\usepackage[utf8]{inputenc} % allow utf-8 input
\usepackage[T1]{fontenc}    % use 8-bit T1 fonts
\usepackage{hyperref}       % hyperlinks
\usepackage{url}            % simple URL typesetting
\usepackage{booktabs}       % professional-quality tables
\usepackage{amsfonts}       % blackboard math symbols
\usepackage{amsmath}
\usepackage{nicefrac}       % compact symbols for 1/2, etc.
\usepackage{microtype}      % microtypography
\usepackage{xcolor}         % colors

\usepackage{enumitem}

\usepackage{graphicx}

\usepackage{multirow}
\usepackage{subfigure}

\usepackage{soul}

\newcommand{\modelname}{\textbf{\texttt{MuTEC}}}
\newcommand{\subone}{\textbf{\texttt{MuTEC\textsubscript{CSE}}}}
\newcommand{\subtwo}{\textbf{\texttt{MuTEC\textsubscript{CEE}}}}
\newcommand{\overall}{\textbf{\texttt{MuTEC\textsubscript{E2E}}}}

\title{Multi-Task Learning Framework for Extracting Emotion Cause Span and Entailment in Conversations}

\author{
Ashwani Bhat \\
Indian Institute of Technology Kanpur (IIT-K) \\
Kanpur, India \\
\texttt{ashubhat44@gmail.com} \\
\And
Ashutosh Modi \\
Indian Institute of Technology Kanpur (IIT-K) \\
Kanpur, India \\
\texttt{ashutoshm@cse.iitk.ac.in} \\
}

\begin{document}

\maketitle

\begin{abstract}
Predicting emotions expressed in text is a well-studied problem in the NLP community. Recently there has been active research in extracting the cause of an emotion expressed in text. Most of the previous work has done causal emotion entailment in documents. In this work, we propose neural models to extract emotion cause span and entailment in conversations. For learning such models, we use RECCON dataset, which is annotated with cause spans at the utterance level. In particular, we propose \modelname,\ an end-to-end Multi-Task learning framework for extracting emotions, emotion cause, and entailment in conversations. This is in contrast to existing baseline models that use ground truth emotions to extract the cause. \modelname\ performs better than the baselines for most of the data folds provided in the dataset.
\end{abstract}

%achieves better cause extraction scores for most of the data folds provided in the dataset.
\section{Introduction}
\vspace{-3mm}

Emotions are an inherent part of human behavior. The choices and actions we make/take are directly influenced by the emotions we are experiencing at any particular moment. Emotions are indicative of and influence the underlying thought process \citep{minsky2007emotion}. Recent developments in AI have made machines an integral part of our lives. For seamless interaction with humans, it is imperative that AI systems understand the emotion experienced by a person and what are the causes and effects of such emotions \cite{singh-etal-2021-end}. Towards this goal in the past two decades, there has been significant research and progress in the area of emotion recognition \cite{9330790}. To understand what influences/causes emotions and how the emotions of a person in turn influence others, recently, there has been active interest in the task of emotion cause extraction (ECE) in documents (\S \ref{sec:related}). %\cite{xia2019emotion, song2020end, ding-etal-2020-ecpe, chen-etal-2020-unified, fan2020transition, ding2020end, wei-etal-2020-effective}. 
\citet{poria2020recognizing} have extended the task of emotion cause extraction to conversations by introducing a new task that requires extraction of the cause span corresponding to a given emotion utterance in a dialogue. The authors have released the \textbf{RECCON} (Recognizing Emotion Cause in CONversations) dataset, where conversations from DailyDialog \cite{li-etal-2017-dailydialog} and IEMOCAP \cite{busso2008iemocap} datasets are annotated with cause span of the emotion utterance. 
\begin{figure}[t]
\centering
\includegraphics[scale=0.22]{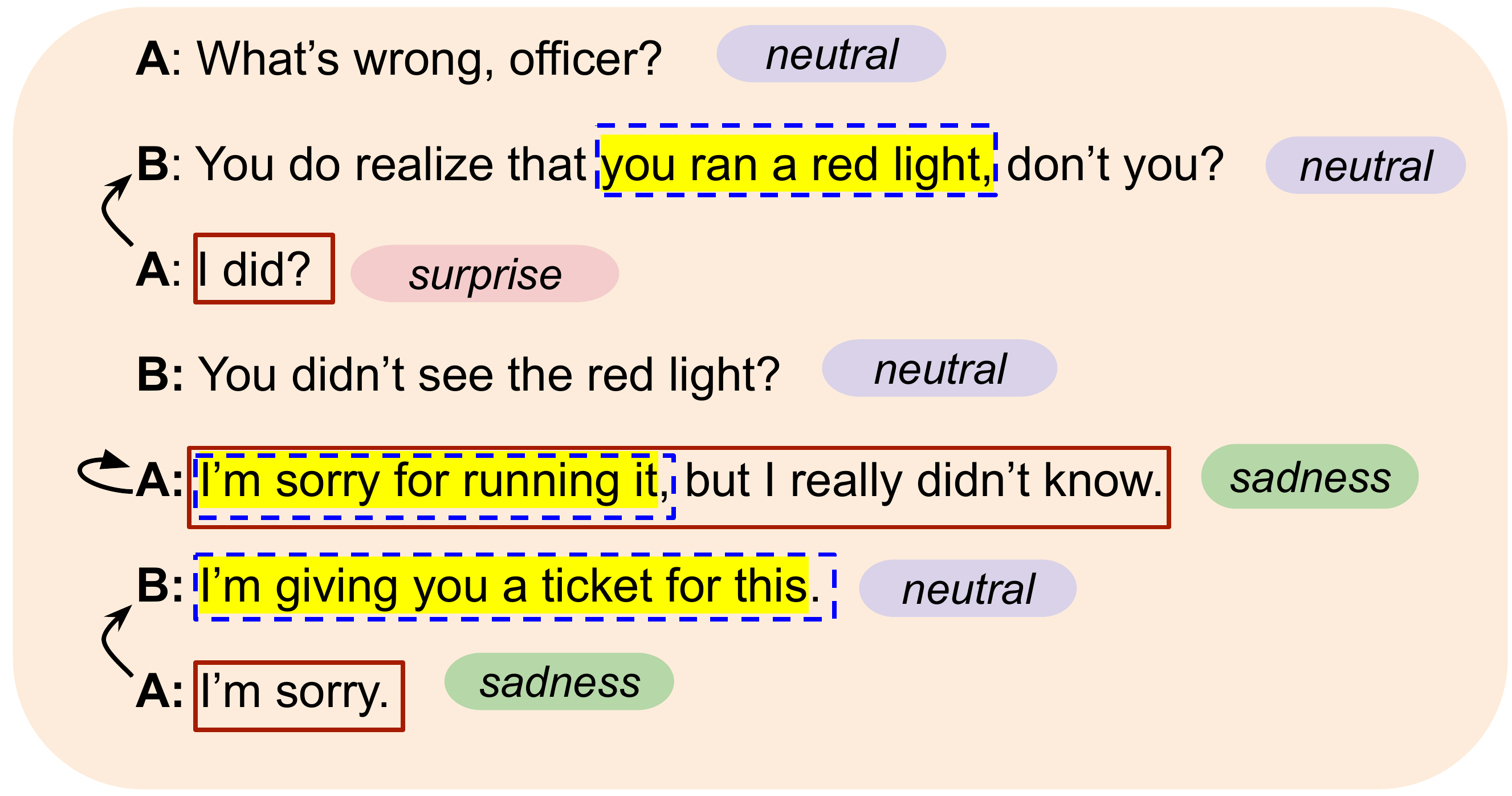}
\caption{Example conversational dialogue from Dailydialog. Each utterance is provided with its corresponding emotion. The dashed rectangles represent the cause spans, and the arrow indicate the cause utterance for any target utterance (solid rectangle) under consideration. As shown, cause for an emotion can sometimes be within the same utterance.}
\label{fig:example}
\vspace{-6mm}
\end{figure}
Fig. \ref{fig:example} shows a sample conversational example showing the emotion cause. The highlighted portion of text represents the cause and the directed arrow A$\rightarrow$B, represents that B contains the cause of A and hence is the cause utterance of A. %For example, \textit{I did ?} has a cause span of \textit{you ran a red light} in the preceding utterance. 
\citet{poria2020recognizing} have introduced two challenging task on RECCON: \textit{Causal Span Extraction} (CSE) and \textit{Causal Emotion Entailment} (CEE) (\S \ref{sec:problem}). The authors used gold emotion annotations during inference. However, this is not a practical assumption. To address this, in this work, we make the following contributions: 
\begin{itemize}[noitemsep,topsep=0pt]
\item For CSE and CEE tasks, we propose an end-to-end Multi-Task learning framework for extracting emotions, emotion cause and cause entailment in conversations (\modelname), where emotions are predicted as auxiliary task and cause span prediction and entailment are the main tasks. We also propose an overall end-to-end model architecture to solve both the tasks using a single architecture. Incorporating emotion prediction directly into the model gives comparable, and in some cases, better performance than models that explicitly use gold emotion labels. We release the code for all the models and experiments: \url{https://github.com/Exploration-Lab/MuTEC}
%\item Incorporating emotion prediction directly into the model gives comparable, and in some cases, better performance than models that explicitly use gold emotion labels.
\item We perform a thorough analysis of the dataset and the models (\S\ref{sec:analysis}). The original RECCON dataset is highly unbalanced with respect to the negative samples resulting in degradation in performance. We create a new version of a balanced dataset and perform experiments on it to show that reducing the negative samples helps to improve the model performance.  
\end{itemize}

\section{Related Work} \label{sec:related}
\vspace{-3mm}

Emotion prediction \cite{joshi-etal-2022-cogmen,bansal-etal-2022-shapes,gargiEmoDef} and emotion generation \cite{colombo-etal-2019-affect,goswamy-etal-2020-adapting} are active areas of research. Emotion Cause Extraction (ECE) \cite{chen2010emotion} is the problem of extracting cause of an emotion given emotion annotations. %at the test time. 
ECE task has attracted significant attention due to its wide applicability. ECE problem has been solved using classical machine learning-based methods \cite{gui-etal-2016-event}, rule-based methods \cite{russo-etal-2011-emocause, neviarouskaya2013extracting}, and deep learning methods \cite{cheng2017emotion, xia2019rthn, li2019context, li2018co, xiao2019context, xu2019extracting, singh-etal-2021-end}. However, the requirement of having gold emotion annotations at the test time has limited its usability in practical scenarios. \citet{li2018co} experimented with removing the annotated emotion, but it led to significant performance drop for the ECE task. Another limitation of ECE task is that it is a two-step process. It first requires annotating the emotions and then extracting their cause, thus ignoring the mutual dependencies between the cause and the emotion. %however this two step process ignores the mutual dependencies between the cause and emotion.

To overcome the limitations of the ECE task, a new task was introduced by \citet{xia2019emotion}: Emotion Cause Pair Extraction (ECPE). This is a more challenging task aimed at  extracting all cause and emotion pairs from the document. ECPE was introduced as a sentence pair classification task. ECPE task doesn't need emotion annotations to be provided at the test time. Also, since it extracts pair of emotion-cause, both the clauses are mutually indicative. To address this task, the authors proposed a two-step approach where they extracted the set of emotions and cause clauses individually in the first step and in the subsequent step, pair and filter the extracted clauses. % based on the outputs in the second. 
This two-step approach suffers from 2 specific problems: (1) The errors from Step 1 are propagated to Step 2 and affect the performance of Step 2. %Hence if Step 1 predicted way too many incorrect emotion or cause clauses, the precision of Step 2 will eventually be reduced. 
(2) The training of the model is not directly aimed at extracting the final emotion-cause clause pair. To address the above issues, a need for an end-to-end architecture was realized. %, and this led to a lot of researchers being interested in this field. 
The first set of work for an end-to-end architecture was done by \citet{ding-etal-2020-ecpe}, where the authors used a representation scheme (in 2D) to represent emotion cause clause pairs and then integrated the cause and emotion pair interaction, prediction, and representation into a single combined framework. 
% Their architecture was named ECPE-2D, where they used a 2D Transformer framework to capture the relationship between different cause and emotion pairs by conducting a simple binary classification on each emotion-cause pair.
\citet{song2020end} and \citet{fan2020transition} solved this problem using a graph-based approach to recognize emotions and their corresponding causes.
% incorporated link prediction from graph representational learning to predict potential edges between unconnected vertices where the edges are directed from the emotion clause to the cause clause. 
% \cite{fan2020transition} framed a transition-based model to convert the ECPE problem into directed graph construction where the model constructs a directed graph and recognizes emotion and corresponding cause simultaneously.
\citet{chen-etal-2020-unified} described this problem as a unified sequence labeling problem, where they extract emotion cause pairs using CNNs. 
% to encode neighboring information, and LSTMs were used for the auxiliary tasks of cause and emotion prediction. 
In \citet{ding2020end}, the authors proposed a multi-label learning framework that extracted both the cause and emotion clauses where the windows for learning multiple labels is fixed on specific cause or emotion clause, and as the position of the clauses is moved, the window also slides.
% \cite{wei-etal-2020-effective} deduced that information about relative position between two clauses is an effective feature for extracting the cause of the emotion. Thus 
\citet{wei-etal-2020-effective} used a ranking strategy where they ranked the emotion-cause clause pair candidates in a given document and modeled this inter-clause relationship using Graph Attention Network \cite{velivckovic2017graph} to perform end-to-end pair extraction.
\citet{singh-etal-2021-end} modeled the mutual interdependence between emotion clause and cause clause using neural networks and trained the entire NN in an end-to-end fashion. 
Recently, \citet{sun2021dual} argued the importance of context in order to extract emotion clause and cause clause and hence proposed a context-aware dual questioning attention network. %that modeled the relationship between emotion-cause clauses using the contextual information.
\citet{ding2020experimental} studied the effect of position bias on Emotion Cause Extraction. Another similar task, Emotion-Cause Span-Pair Classification and Extraction was proposed by \citet{bi2020ecsp}, in which instead of taking a definite emotion and cause clause, they took random spans of text from the document that may span across multiple clauses.\\
Recently, emotion classification in conversations has been an active research area. \citet{wang2020contextualized, shen2020dialogxl, chapuis2020hierarchical} use transformer-based architectures to recognize emotions. \citet{sheng2020summarize, ghosal2019dialoguegcn, zhong2019knowledge} use graph neural networks and sequence-based networks to model the relationship between utterances and recognize the emotions. \citet{ishiwatari2020relation} use both contextual embeddings from transformer-based models and graph neural networks to recognize the emotions. %\citet{poria2020recognizing} introduce a new task of \textit{Recognizing Emotion Cause in CONversations} along with an utterance level cause span annotated dataset: \textbf{RECCON}. %A subset of DailyDialog and IEMOCAP  datasets are annotated with the utterance level cause span. %The focus of this paper is the RECCON dataset. 

%RECCON introduced two tasks of Cause Span Extraction and Causal Emotion Entailment. A three-fold dataset was created using RECCON consisting of both positive and negative samples, along with transformer-based baselines on Roberta \cite{liu2019roberta} and SpanBERT \cite{joshi2020spanbert}. They used emotion annotations during inference for both the given tasks. \AB{done}
%\AM{maybe write 1-2 more lines about this.}
%As described in the next section, they introduced two challenging tasks on the RECCON dataset. 
\section{RECCON Tasks}\label{sec:problem}
\vspace{-3mm}

RECCON dataset introduces two tasks for extracting emotion cause in conversations:

\noindent\textbf{Task 1: Causal Span Extraction:} This task involves finding the emotional cause span for a target utterance. The task has two settings. (1) In the first setting,  \textit{conversational history} is not considered (\textbf{(w/o CC)}). (2) In the other setting \textit{conversational history} is considered (\textbf{(w/ CC)}). 

\noindent\textbf{Task 2: Causal Emotion Entailment:} This task involves determining whether the candidate utterance causally entails the emotion utterance or not. This task is also formulated in two settings. (1) without Conversational Context \textbf{(w/o CC)}. (2) with Conversational Context \textbf{(w/ CC)}. 

Three fold dataset (\S\ref{sec:experiments}) is created using RECCON consisting of both positive and negative samples, RoBERTa \cite{liu2019roberta} and SpanBERT \cite{joshi2020spanbert} are used as the prediction models. \citet{poria2020recognizing} use gold emotion annotations during inference for both the given tasks. The authors solve \textbf{Cause Span Extraction} as a SQuAD like question answering task where target and cause utterance form the question and the answer contains the cause span. 
For Fig. \ref{fig:example}, a positive sample is created as: \textit{\textbf{Context:}} \textit{``What's wrong, officer? You do realize that you ran a red light, don't you? I did?"}
\textit{\textbf{Question:}} \textit{``The target utterance is I did. The evidence utterance is You do realize that you ran a red light, don't you. What is the causal span from evidence in the context that is relevant to the target utterance’s emotion Surprise?"}
\textit{\textbf{Answer:}} \textit{``you ran a red light"}.  Here, the task is to predict the answer span from the context for a given question.\\
\textbf{Causal Emotion Entailment} is solved as Natural Language Inference (NLI) task. For solving this task, a binary labelled dataset is created as \textit{<Context> <SEP> <Utterance> <SEP> <Candidate Cause utterance> <SEP> <History>}. For example, a positive sample for Fig \ref{fig:example} is created as: \textit{``surprise $<SEP>$ I did? $<SEP>$ you do realize that you ran a red light, don't you? $<SEP>$ What's wrong, officer? You do realize that you ran a red light, don't you? I did?"}. Here, the task is to predict a binary entailment label of 0 if candidate cause utterance doesn't contain the cause of given utterance and a label of 1 if candidate cause utterance contains the cause for a given utterance. 

However, we approach the problem differently. For both the tasks CSE and CEE, the input to the model is concatenation of target utterance, cause utterance and context (more details about the  dataset in App. \ref{app:reccon_tasks}). For example, for Fig. \ref{fig:example} the corresponding input to the model is:
\textit{I did?. you do realize that you ran a red light, don't you? $<SEP>$ What's wrong, officer? You do realize that you ran a red light, don't you? I did?" 
}. Given the input in this format, for CSE, the task is to predict start and end positions in the context. For CEE, the task is to predict entailment label as 1 or 0. %The dataset format is described in more detail in Appendix \ref{app:reccon_tasks}.

%For our case, we are constructing same dataset for both the tasks. For example, a positive sample in Fig. \ref{fig:example} will be like:
%\textit{I did?. you do realize that you ran a red light, don't you? $<SEP>$ What's wrong, officer? You do realize that you ran a red light, don't you? I did?" }. For CSE, the loss will be calculated based on start and end positions in the context. For CEE, we use entailment label, for positive sample as 1 and negative sample as 0. The dataset format is further discussed in Appendix \ref{app:reccon_tasks}.

%\AM{need to briefly describe how Poria solved these tasks by posing it as a machine comprehension problem, and more details can be provided in the appendix}

% \subsection{Task 1: Causal Span Extraction}
% \textbf{Causal Span Extraction} is a problem in determining the emotional cause span of a target utterance. There are two settings for this task. (1) In the first setting,  \textit{conversational history} is not being considered (\textbf{without Conversational Context (w/o CC)}). (2) In the other setting \textit{conversational history} is being considered (\textbf{with Conversational Context (w/ CC)}).

% \subsection{Task 2: Causal Emotion Entailment}
% \textbf{Causal Emotion Entailment} is a problem of determining whether the given candidate utterance is the cause utterance for the given emotion utterance. This task is also formulated in two settings. (1) \textbf{without Conversational Context (w/o CC)}. (2) \textbf{with Conversational Context (w/ CC)}. 
\section{Proposed Models} 
\label{sec:models}
\vspace{-3mm}
% ----------- twostep architecture (sub1) -----------------
% \begin{figure}[t]
% \centering
%   \includegraphics[scale=0.35]{TL4NLP/images/two_step_2d.pdf}
% %  \includegraphics[width=0.6\linewidth]{TL4NLP/images/two_step_2d.pdf}
%   \caption{Two step architecture for Cause Span Extraction. The first step is an \textit{Emotion Predictor} network followed by the \textit{Cause Span Predictor} network (second step).}
%   \label{fig:two_step}
% \end{figure}

We develop transformer-based multi-task learning models that learn both emotion and emotion cause without being provided with the gold emotion annotations during inference. We perform transfer learning via pre-trained transformer-based LMs. 
%\AB{Done}
%\AM{Give a brief intro of a few lines, kind of an outline. What models we develop and how is diff, etc.}
%The following are the proposed models for both the emotion cause tasks.

\subsection{Task 1: Cause Span Extraction} 
\vspace{-3mm}
% We propose two models for solving this task: a two-step architecture and an end-to-end architecture. 
As an initial baseline, we solved this problem using two-step model consisting of an \textit{Emotion Predictor (EP)} followed by the \textit{Cause-Span Predictor (CSP)}. An advantage of the two-step model is that it is modular; separate  architectures can be applied for both the emotion predictor and cause span predictor. However, there are two drawbacks: 1) The error in the first step is propagated to the next step, and 2) Such an approach assumes that emotion prediction and cause-span prediction are mutually exclusive tasks. %, which is not true in general. 
To overcome these limitations, we propose an end-to-end architecture.

% ----------- end2end architecture -----------------

\begin{figure*}[t]%
\centering
\label{fig:subone_fig}%
\subfigure[Training architecture for \subone]{%
\label{fig:one_step_train}%
\includegraphics[scale=0.25]{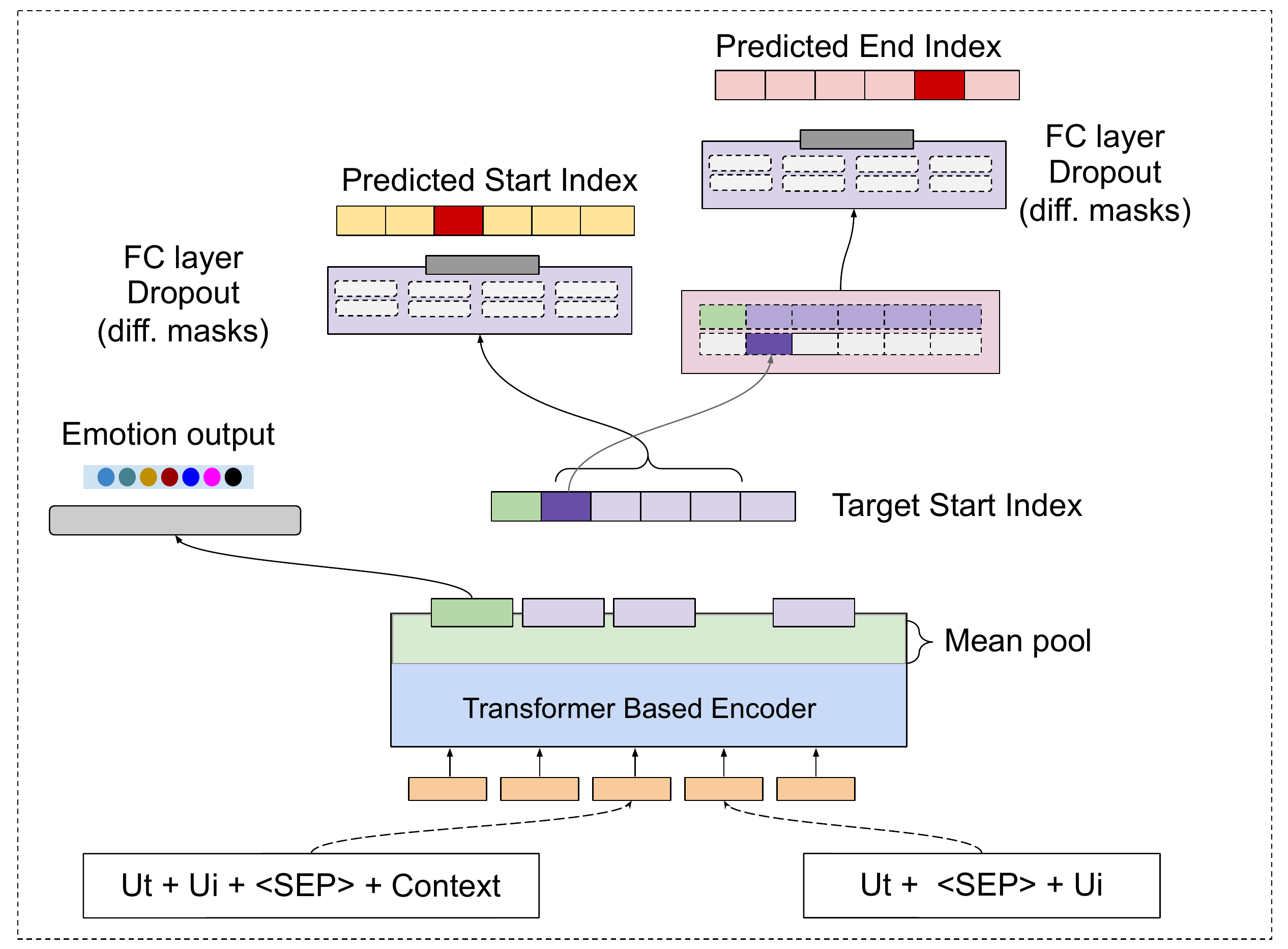}}%
\qquad
\subfigure[For inference, output is $k \times k$ start-end index pairs.]{%
\label{fig:one_step_inference}%
\includegraphics[scale=0.25]{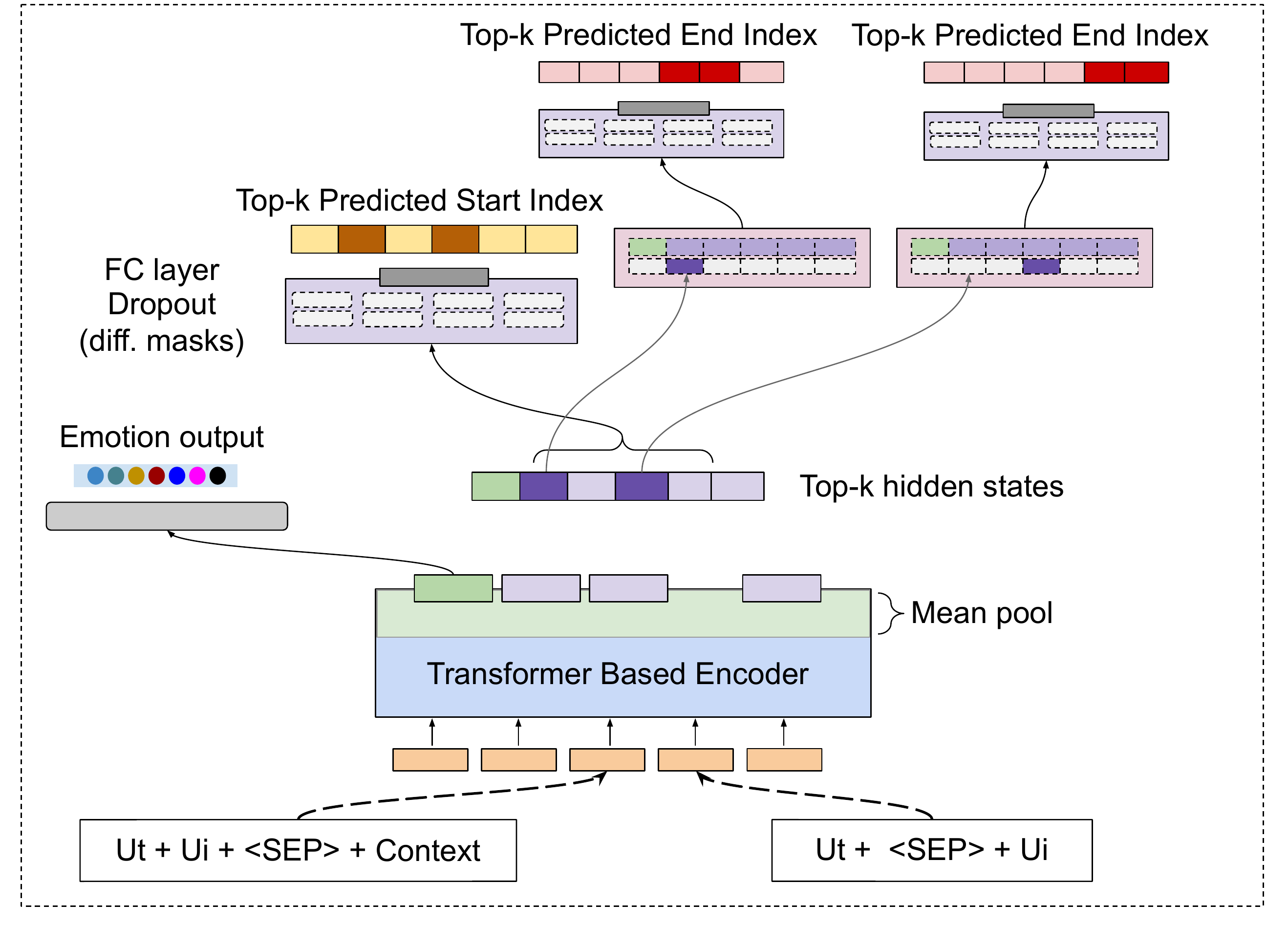}}%
\vspace{-4mm}
\caption{End-to-End architecture (\subone) for Cause Span Extraction, for training and inference.}
\vspace{-3mm}
\end{figure*}

\noindent\textbf{End-to-End Architecture (\subone):} \subone\ is an end-to-end multi-task framework where we perform cause span extraction as the main task and emotion prediction
as an auxiliary task (Fig.\ref{fig:one_step_train}). %The architecture is shown in Fig.\ref{fig:one_step_train}. 
For inference, we use a slightly modified version of the training architecture (Fig.\ref{fig:one_step_inference}). The architecture is inspired from the idea that lets the model use the start position information to predict the end token position.

\noindent\textbf{\subone Training:} The input consists of target utterance $U_t$, candidate cause utterance $U_i$ and $Context$ (w/ CC setting) or no context (w/o CC setting). The input is passed into a transformer based pre-trained model, we mean pool all the 12 layers of pre-trained model to get sequence output: $\mathrm{pool, h_1, h_2,\ldots} = E_{sb}(U_t U_i \texttt{<SEP>} \textit{Context})$ and $\mathrm{pool', h_1', h_2',\ldots} = \mathrm{meanpool}(pool, h_1, h_2,..)_{12}$. The pooled output is used to predict the auxiliary task of emotion prediction. It is passed through a MLP layer and then through a softmax to get the predicted emotion: $ emotion^{logit} = \mathrm{MLP}(\mathrm{pool'})$, and $emotion_{pred} = \sigma(emotion^{logit})$. During training, the given start position is used to predict the end index. The start hidden state ($h_s$) and the original hidden states ($h_1'$, $h_2', \ldots$) are concatenated. The  concatenated hidden states are passed through a multi sample dropout (MSDropout) \cite{inoue2019multi} to get the predicted end logit. This end logit is then passed through a softmax layer: $start^{logit} = \mathrm{MSDropout}([h_1', h_2', \ldots])$; $start_{pred} = \sigma(start^{logit})$, $end^{logit} = \mathrm{MSDropout}([h_1', h_2', \ldots]\oplus h_s)$, and $end_{pred} = \sigma(end^{logit})$. Here, $\oplus$ is the concatenation operation. The training loss is a linear combination of the loss for cause-span prediction and emotion prediction: $\mathcal{L}_{total} = \mathcal{L}_{cause\_span} + \mathbf{\beta}\mathcal{L}_{emotion}$. $\beta$ is a hyperparameter, determined using the validation set.
%  ----------- end2end architecture (train) -----------------
% \begin{figure}[!ht]
% \centering
%  \includegraphics[scale=0.3]{TL4NLP/images/final_one_step_train_2d.pdf}
%  % \includegraphics[width=0.6\linewidth]{TL4NLP/images/final_one_step_train_2d.pdf}
%   \caption{\textbf{End-to-End Cause Span Extraction (training)}}
%   \label{fig:one_step_train}
% \end{figure}
% ----------- end2end architecture (inference) -----------------
% \begin{figure}[!ht]
% \centering
% \includegraphics[scale=0.3]{TL4NLP/images/final_one_step_inference_2d.pdf}
%   %\includegraphics[width=0.6\linewidth]{TL4NLP/images/final_one_step_inference_2d.pdf}
%   \caption{\textbf{E2E Cause Span Extraction (inference)}. The output will be $k \times k$ start-end index pairs. An $\mathrm{argmax}$ is taken over the start-end pairs.}
%   \label{fig:one_step_inference}
% \end{figure}

{\label{sec:mutec_infer}
\noindent\textbf{\subone\ Inference:} During inference, we are not provided with start positions. Hence we find top-$k$ start indices and concatenate the hidden state of each such index to original hidden states, thus creating $k$ different end candidate logits. For each of such $k$ end logits, we again find top-$k$ end indices. We refer this $k$ as the \textit{beam size.} This creates $k \times k$ start-end index pairs, and \texttt{argmax} over these $k \times k$ gives the predicted start and end index.}

\subsection{Task 2: Causal Emotion Entailment}
\begin{figure*}[t]%
\centering
\label{fig:subone_fig}%
\subfigure[End-to-End architecture (\subtwo) for Causal Emotion Entailment]{%
\label{fig:entail_e2e}%
\includegraphics[scale=0.30]{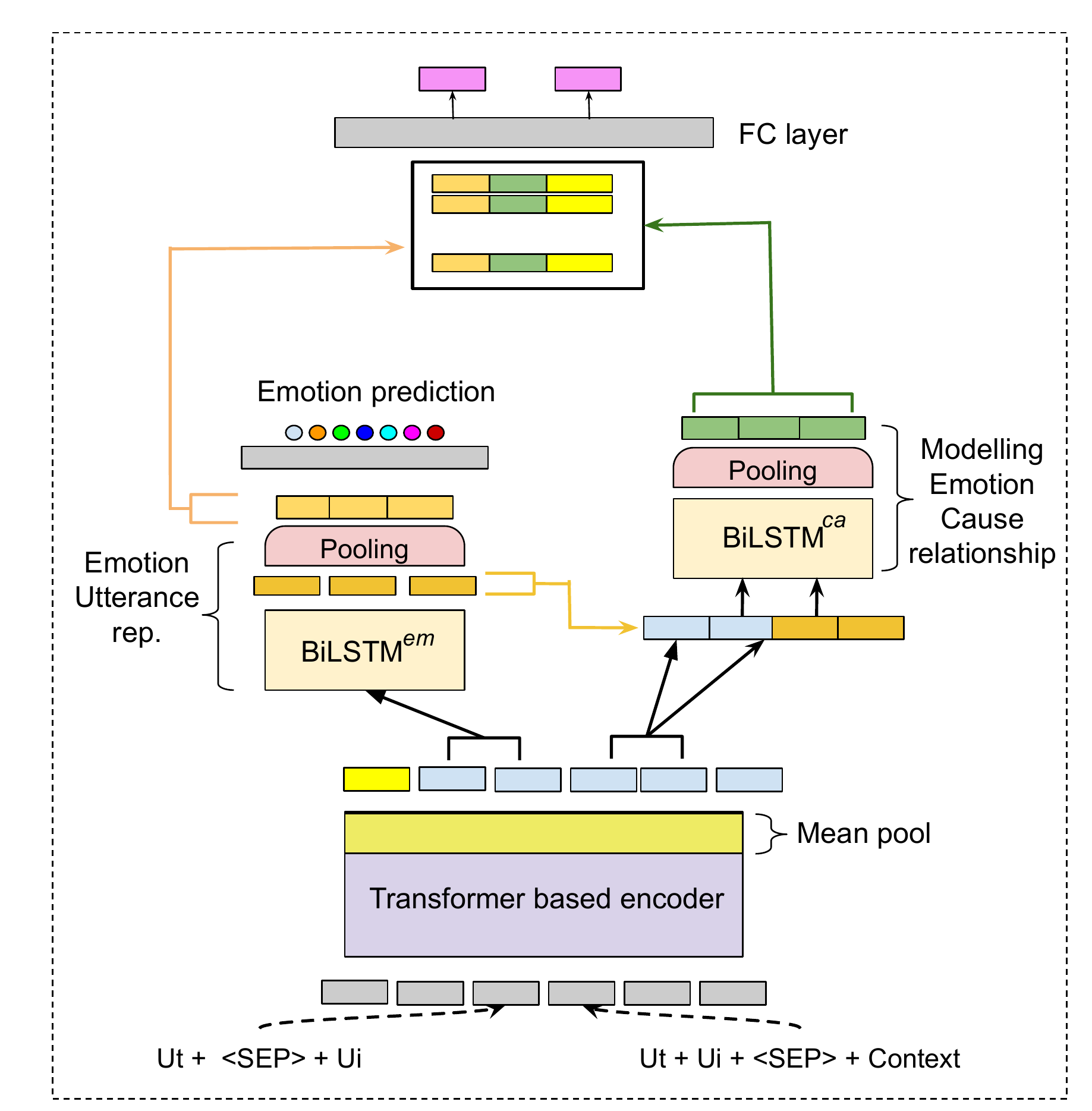}}%
\qquad
\subfigure[E2E model for CSE and CEE (\overall)]{%
\label{fig:overall_e2e}%
\includegraphics[scale=0.30]{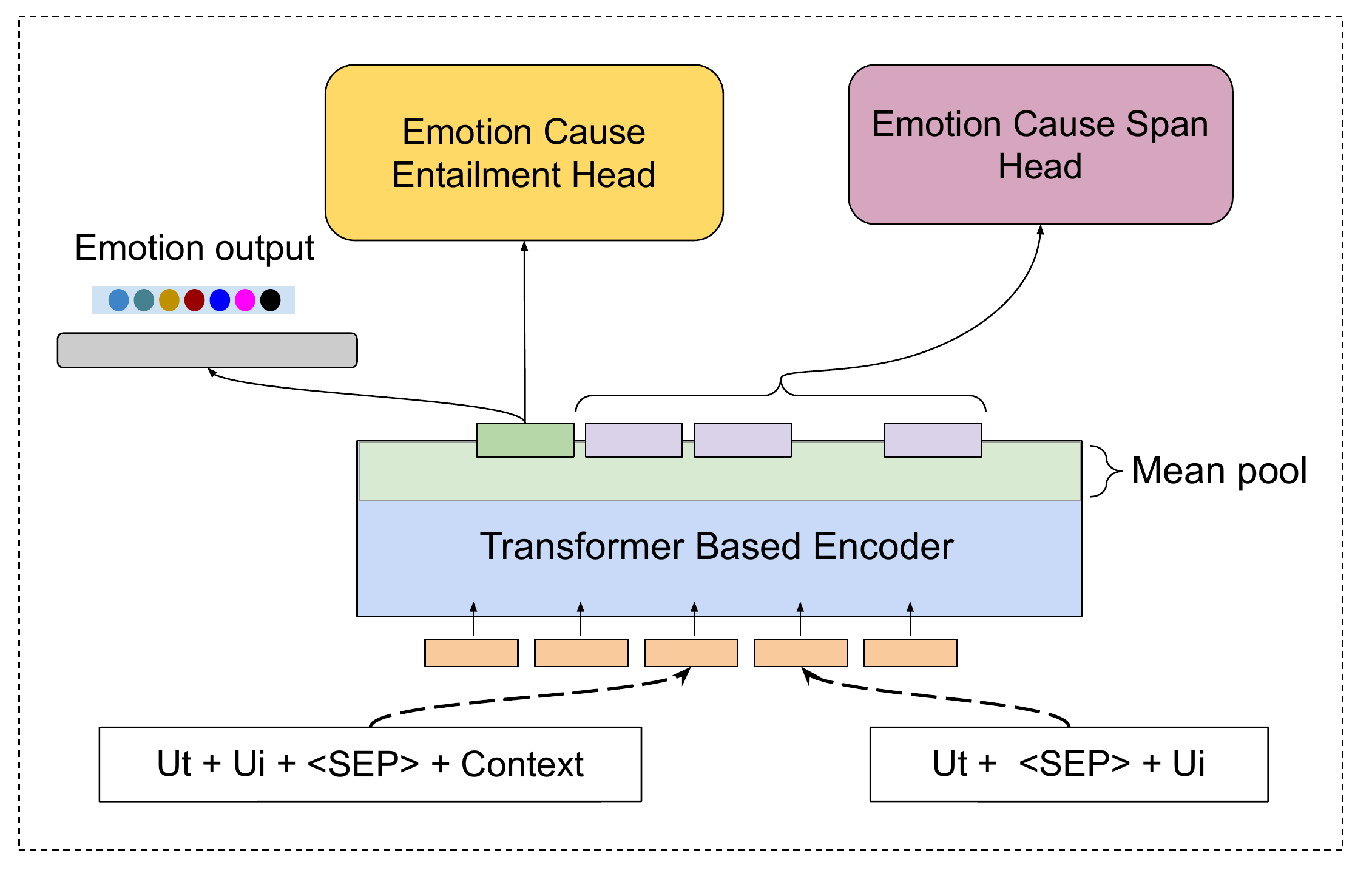}}%
\vspace{-3mm}
\caption{Proposed Models}
\vspace{-3mm}
\end{figure*}

\vspace{-3mm}
For the task of Causal Emotion Entailment, we propose a multi-task learning approach, \subtwo, that consists of three components (Fig. \ref{fig:entail_e2e}). The first component learns contextual representations of the input, i.e., target utterance, candidate cause utterance, and the context. Second component models the relationship between cause and emotion utterances to obtain better representations. Finally, the third component concatenates all the representations and performs entailment (a sentence pair classification task). In order to learn better emotion representations, we include emotion prediction as an auxiliary task. %The overall end-to-end architecture for Causal Emotion Entailment is shown in Figure \ref{fig:entail_e2e}.

% \begin{figure}[!ht]
% \centering
% \includegraphics[scale=0.30]{TL4NLP/images/entail_end2end_2d.pdf}
%   %\includegraphics[width=0.6\linewidth]{TL4NLP/images/entail_end2end_2d.pdf}
%   \caption{End-to-End architecture (\subtwo) for Causal Emotion Entailment}
%   \label{fig:entail_e2e}
%   \vspace{-4mm}
% \end{figure}

% ----- testing ----

% ------- testing ----

\noindent\textbf{Learning Contextual Representations:}
Given an input: $U_t + U_i+\texttt{<SEP>}+ Context$ (w/ CC) and $U_t+\texttt{<SEP>}+U_i$ (w/o CC), we use the RoBERTa model to encode the input and learn contextualized representations: $ pool, h1, h2,..=E_{rb}(U_t U_j \texttt{<SEP>} Context)$.  We empirically found out that mean-pooling last $4$ hidden layers gave the best results. 

\noindent\textbf{Modelling Emotion-Cause relationship:}
The representations of $U_t$ from the first component are then passed into a first-token-level BiLSTM ($BiLSTM^{em}$), for capturing only the target utterance's context and to predict utterance emotions (by mean pooling the representations and passing it through a single layer neural network): $[h^{j'}_{t_1}, h^{j'}_{t_2}, h^{j'}_{t_3}, ...] =BiLSTM^{em}([h^j_{t_1}, h^j_{t_2}, h^j_{t_3}, ...])$. For the auxiliary emotion prediction, the output is mean-pooled and passed through a single neural network: $H_{jt}=\mathrm{meanpool}([h^{j'}_{t_1}, h^{j'}_{t_2}, h^{j'}_{t_3}, ...])$ and $y'_{aux}=\sigma(MLP(H_{jt})$. The hidden representations from $BiLSTM^{em}$ i.e., ${(h^{j'}_{t_1}, h^{j'}_{t_2}, h^{j'}_{t_3}, ...)}$ are first concatenated with the representations of $U_i$ and then passed into another BiLSTM ($BiLSTM^{ca}$) that uses target utterance's representations to capture the cause utterance. This is used to model the relationship between cause and emotion utterance. The output of $BiLSTM^{ca}$ gives utterance's emotion-cause representations: $[h^{j''}_1, h^{j''}_2, \ldots] = BiLSTM^{ca}([h^{j'}_1, h^{j'}_2, \ldots])$, where, $[h^{j'}_1, h^{j'}_2, \ldots] = [h^{j'}_{t_1}, h^{j'}_{t_2}, \ldots] \oplus [h^j_{i_1}, h^j_{i_2}, \ldots]$.

% These are represented as ${(h^{j''}_1, h^{j''}_2, h^{j''}_3, \ldots)}$.
% \begin{align*}
% [h^{j'}_1, h^{j'}_2, \ldots] &= [h^{j'}_{t_1}, h^{j'}_{t_2}, \ldots] \oplus [h^j_{i_1}, h^j_{i_2}, \ldots]\\
%     [h^{j''}_1, h^{j''}_2, \ldots] &= BiLSTM^{ca}([h^{j'}_1, h^{j'}_2, \ldots])
% \end{align*}
%where $\oplus$ is the concatenation operation.

\noindent\textbf{Entailment Prediction:} The pooled representation from $BiLSTM^{em}$ and $BiLSTM^{ca}$ are concatenated along with the pool output of the pre-trained encoder ($[pool]$) and then passed through a simple MLP to perform classification: $y' = \sigma(MLP(\mathbf{X}_j))$. Here, $\mathbf{X}_j = [pool\oplus H_{jt} \oplus H_{ji}]$ and $H_{ji} = \mathrm{meanpool}([h^{j''}_1, h^{j''}_2, h^{j''}_3, \ldots])$, $H_{jt} = \mathrm{meanpool}([h^{j'}_{t_1}, h^{j'}_{t_2}, h^{j'}_{t_3}, \ldots])$.
% \begin{align*}
% H_{jt} &= \mathrm{meanpool}([h^{j'}_{t_1}, h^{j'}_{t_2}, h^{j'}_{t_3}, \ldots])\\
% H_{ji} &= \mathrm{meanpool}([h^{j''}_1, h^{j''}_2, h^{j''}_3, \ldots])\\
% \mathbf{X}_j &= [pool\oplus H_{jt} \oplus H_{ji}]\\
% y' &= \sigma(MLP(\mathbf{X}_j))    
% \end{align*}

%where $\oplus$ is a concatenation operation and $\sigma$ is a softmax function.

\noindent\textbf{Loss Function:} The combined loss function is given by $\mathcal{L}_{total} = \mathcal{L}_{entail} + \mathbf{\beta}\mathcal{L}_{emotion}$. Validation set is used to determine $\beta$. Since the dataset is highly unbalanced, we use weighted cross-entropy loss for both, where the weights are distributed as inverse of the number of class instances.
\subsection{E2E Cause Span and Entailment model}

In order to perform the end to end training for both the tasks using a single model, we used a similar architecture to Fig. \ref{fig:one_step_train} and Fig. \ref{fig:one_step_inference} and added a cause entailment head on top (Fig. \ref{fig:overall_e2e}). %describes this end to end architecture.
The model uses a transformer based encoder (RoBERTa-base) as base layer with three heads, namely, emotion head, cause span head and cause entailment head, on top. The entire model is trained end-to-end. The emotion and cause entailment head is single layered neural net. The emotion cause span head is similar to what is used in \subone. %, as shown in Fig.  \ref{fig:one_step_train}.

\noindent\textbf{Loss Function:} The overall loss function for the end to end model is: $\mathcal{L}_{total} = \mathcal{L}_{causespan} + \mathcal{L}_{entail} + \mathcal{L}_{emotion}$. Since the dataset is highly unbalanced, we use weighted cross-entropy loss, where the weights are distributed as inverse of the number of class instances.
\section{Experiments and Results} \label{sec:experiments}
\vspace{-3mm}
%\AM{Describe the baseline models (basically Poria's method) and folds, our experiments.}
%\AB{done}

% \begin{figure}[t]
% \centering
% \includegraphics[scale=0.25]{TL4NLP/images/overall_end2end.pdf}
%   \caption{E2E model for CSE and CEE (\overall).}
%   \label{fig:overall_e2e}
%   \vspace{-2mm}
% \end{figure}

% ----------- Task 1 ------------------ %
\begin{table*}[t]
\renewcommand{\arraystretch}{1.5}
\setlength\tabcolsep{15pt}
\resizebox{\columnwidth}{!}{
\begin{tabular}{|c|c|c|c|c|c|c|c|c|c|c|c|c|}
\hline
\multirow{2}{*}{\textbf{Train Fold}} &
  \multirow{2}{*}{\textbf{Test Fold}} &
  \multirow{2}{*}{\textbf{Model}} &
  \multicolumn{5}{c|}{\textbf{w/o CC}} &
  \multicolumn{5}{c|}{\textbf{w/ CC}} \\ \cline{4-13} 
 &
   &
   &
  Emotion Acc. &
  \textbf{$EM_{pos}$} &
  \textbf{$F1_{pos}$} &
  \textbf{$F1_{neg}$} &
  \textbf{$F1$} &
  Emotion Acc.&
  \textbf{$EM_{pos}$} &
  \textbf{$F1_{pos}$} &
  \textbf{$F1_{neg}$} &
  \textbf{$F1$} \\ \hline 
\multirow{30}{*}{Fold1 (DD)} &
  \multirow{5}{*}{Fold1 (DD)} &
  RoBERTa-base &
- & 26.82 & 45.99 & \textbf{84.55} & \textbf{73.82} & - & 31.63 & 58.17 & 85.85 & \textbf{75.45} \\ \cline{3-13} 
&  & 
SpanBERT &
- & 33.26 & 57.03 & 80.03 & 69.78 & - & 34.64 & 60.00 & \textbf{86.02} & 75.71 \\ \cline{3-13} 
& & Two Step &      
80.43 & 32.11 & 53.44 & 81.87 & 67.54 & 82.12 & 34.22 & 58.90 & 83.12 & 72.13 \\ \cline{3-13} 
& & \subone &      
82.54 & \textbf{36.29} & \textbf{62.12} & 61.86 & 53.76  & 83.42 & \textbf{36.87} & \textbf{66.92} & 73.89 & 62.90
\\ \cline{3-13} 
& & \overall &      
80.12 & 35.47 & 61.74 & 64.87 & 55.74  & 82.02 & 35.78 & 64.11 & 75.41 & 63.24
\\ 
\cline{2-13} 
& \multirow{5}{*}{Fold2 (DD)}   & 
RoBERTa-base  & 
- & 26.82 & 45.99 & 83.52 & 72.66 & - & 32.95 & 59.02 & 95.36 & 87.63 \\ \cline{3-13} 
& & SpanBERT  & 
- & 33.26 & 57.03 & \textbf{84.02} & \textbf{74.80} & - & 32.37 & 57.04 & 95.01 & 87.00 \\ \cline{3-13} 
& & Two Step &      
81.13 & 32.43 & 54.24 & 83.98 & 73.04 & 76.82 & 34.24 & 61.66 & 94.78 & 87.08 \\ 
\cline{3-13} 
& & \subone &      
81.24 & \textbf{35.94} & \textbf{62.42} & 64.20 & 54.18 & 69.16 & \textbf{36.10} & \textbf{66.04} & \textbf{96.88} & \textbf{89.73} \\ 
\cline{3-13} 
& & \overall &      
75.56 & 35.78 & 61.22 & 64.56 & 53.48 & 72.87 & 35.28 & 64.21 & 95.39 & 88.47
\\ 
\cline{2-13} 
& \multirow{5}{*}{Fold3 (DD)}   &
RoBERTa-base  & 
- & 26.82 & 45.99 & \textbf{81.50} & \textbf{70.26} & - & 32.95 & 59.02 & 95.37 & 87.65 \\ \cline{3-13} 
& & SpanBERT & 
- & 33.26 & 57.03 & 79.65 & 69.83 & - & 32.31 & 56.99 & 94.92 & 86.87 \\ \cline{3-13} 
&  & Two Step  &      
81.13 & 32.43  & 54.24  & 79.90 & 66.88 & 84.14 & 34.24 & 61.66 & 96.44 & 86.80 \\ 
\cline{3-13} 
&  & \subone  &      
80.57 & \textbf{35.94}  & \textbf{62.42}  & 62.10 & 51.06 & 86.12 & \textbf{36.10} & \textbf{66.04} & \textbf{96.85} & \textbf{88.20}\\ 
\cline{3-13} 
& & \overall &      
81.34 & 35.78 & 61.22 & 63.35 & 52.74 & 80.19 & 35.28 & 64.21 & 96.48 & 87.96 

\\ 
\cline{2-13} 
& \multirow{5}{*}{Fold1 (IEMO)} 
& RoBERTa-base  & 
 - & 9.81 & 18.59 & \textbf{93.45} & \textbf{87.60} & - & 10.19 & 26.88 & 91.68 & 84.52 \\ \cline{3-13} 
&  & SpanBERT   & 
- & 16.20 & 30.22 & 87.15 & 77.45 & - & 22.41 & 37.80 & 90.54 & 82.86 \\ \cline{3-13} 
&   & Two Step  & 
23.20 & 18.56 & 34.12 &  86.66 & 74.42  &  22.24   &  20.12 & 33.36 & \textbf{93.62}  &  \textbf{86.72}  \\ 
\cline{3-13} 
&   & \subone  & 
23.66 & \textbf{30.52} & \textbf{50.68} &  70.21 & 55.64  &  21.22 & \textbf{31.60} & \textbf{53.62} & 81.78  &  72.56  \\ 
\cline{3-13} 
& & \overall &      
25.34 & 26.78 & 47.32 & 75.68 & 53.47 & 17.92 & 30.74 & 50.39 & 83.74 & 75.63 \\
\cline{2-13} 
& \multirow{5}{*}{Fold2 (IEMO)} 
& RoBERTa-base  & 
- & 9.81  & 18.59 & \textbf{92.18} & \textbf{85.41} & - & 10.93 & 28.26 & 95.49 & 90.85 \\ \cline{3-13} 
&  & SpanBERT  & 
- & 16.20 & 30.22 & 88.63 & 79.80 & - & 24.07 & 40.57 & 96.28 & 92.41 \\ \cline{3-13} 
&  & Two Step &
26.86 & 18.56 & 34.12 & 87.80  & 76.52 & 28.18  & 23.56 & 35.60 & 94.86  & 91.22  \\ 
\cline{3-13} 
&  & \subone &
25.54 & \textbf{30.52} & \textbf{50.68} & 71.52  &  57.60 & 27.18  & \textbf{30.32} & \textbf{53.62} & \textbf{96.60}  & \textbf{92.96}  
\\ 
\cline{3-13} 
& & \overall &      
28.31 & 26.78 & 47.32 & 76.23 & 56.95 & 15.20 & 30.11 & 52.75 & 96.23 & 92.57 
\\ 
\cline{2-13} 
& \multirow{5}{*}{Fold3 (IEMO)} 
& RoBERTa-base  & 
- & 9.81  & 18.59 & \textbf{91.82} & \textbf{84.83} & - & 10.93 & 28.26 & 95.47 & 90.81 \\ \cline{3-13} 
&  & SpanBERT & 
- & 16.20 & 30.22 & 86.95 & 77.25 & - & 24.07 & 40.57 & 96.28 & 92.41 \\ \cline{3-13} 
&  & Two Step  &
26.86 & 18.56 & 34.12  & 86.60 &  75.84 & 28.18 & 23.56  & 35.60  &  94.80 & 92.40      \\
\cline{3-13} 
&  & \subone  &
27.30 & \textbf{30.52} & \textbf{50.68}  & 72.63 &  58.16 & 25.30 & \textbf{30.32} & \textbf{53.62}  & \textbf{97.96}  & \textbf{94.40}   
\\ 
\cline{3-13} 
& & \overall &      
30.78 & 26.78 & 47.32 & 75.66 & 59.78 & 17.02 & 30.11 & 52.75 & 96.80 & 93.97 
\\ \hline
\end{tabular}
}
%\vspace{0.5mm}
\caption{Results for Cause Span Extraction task for Two Step, \subone and \overall on RECCON-DD and RECCON-IEMO. IEMO dataset is only used in the inference phase.}
\label{tab:task1_twostep}
\vspace{-3mm}
\end{table*}

\noindent\textbf{Dataset:}
%\noindent\textbf{Dataset:} 
The RECCON corpus annotates \textit{DailyDialog} \cite{li-etal-2017-dailydialog} and \textit{IEMOCAP} \cite{busso2008iemocap} corpora with emotion cause information. In order to train a model, the instances which are not the cause of an utterance (non-cause utterances) were used to create negative samples. Three strategies are adopted by RECCON to create negative samples resulting in 3 data folds. \textit{Fold 1:} negative examples are created as ${(U_t,U_i) | U_i\in H(U_t)-C(U_t)}$, here, $H(U_i)$ is the conversational history and $C(U_t)$ is collection of causal utterances for $U_t$. \textit{Fold 2:} any non-causal utterance $U_i$ is selected randomly along with its conversational history $H(U_i)$ from another dialogue to construct negative examples. \textit{Fold 3:} same as Fold 2, but the only constraint here is that the utterance $U_i$ sampled from another dialogue should have the same emotion as the target utterance $U_t$ to create the negative example (details in App. \ref{app:reccon}).

%\subsection{Model Training and Inference}
\noindent\textbf{Model Training and Inference:}
%\noindent\textbf{Model Training and Inference:} %We provide the model hyper-parameters in Appendix \ref{app:hyperparams}. 
For evaluation, the models are trained on one fold, and other folds are used for inference (hyper-parameters in Appendix \ref{app:hyperparams}). IEMO is the annotated IEMOCAP dataset which is only used for inference as the number of samples in the annotated IEMOCAP dataset is less for training. The experimental results are averaged across 3 runs to account for  the variance in transformer based models. 

\noindent\textbf{Cause Span Extraction Task:}
SpanBERT (finetuned on SQUaD) and RoBERTa Base with a linear layer on top is used as the baseline by \citet{poria2020recognizing}. Models are evaluated using Exact match (EM), Positive F1, Negative F1, and Overall F1 (details in App. \ref{app:metrics}). A positive F1 score considers only positive samples (i.e., utterances having cause spans). The negative sample has an empty span. %cause span.
%The baselines for the task of Cause Span Extraction were given as SpanBERT (finetuned on SQUaD) and RoBERTa Base with a linear layer on top. This task was solved as a Question Answering task by the baseline.

% \noindent\textbf{Evaluation Metrics:} For evaluation of models, metrics used are Exact match (EM), Positive F1, Negative F1, and Overall F1. Details provided in Appendix \ref{app:metrics}. 
% A positive F1 score will consider only positive samples (i.e., utterances having cause spans). The negative sample has an empty cause span. 

% \AM{Write briefly about the evaluation metrics used and refer the details to Appendix, i have already moved things there}\AB{done}

% \AM{Write few lines about the experiments ran}

% Please add the following required packages to your document preamble:
% \usepackage{multirow}

\begin{table*}[t]
% \footnotesize
\renewcommand{\arraystretch}{1.5}
\setlength\tabcolsep{15pt}
\resizebox{\columnwidth}{!}{
\begin{tabular}{|c|c|c|c|c|c|c|c|c|c|c|}
\hline
\multirow{2}{*}{\textbf{Train Fold}} &
  \multirow{2}{*}{\textbf{Test Fold}} &
  \multirow{2}{*}{\textbf{Model}} &
  \multicolumn{4}{c|}{\textbf{w/o CC}} &
  \multicolumn{4}{c|}{\textbf{w/ CC}} \\ \cline{4-11} 
 &
   &
   &
  Emotion Acc. &
  \textbf{$F1_{pos}$} &
  \textbf{$F1_{neg}$} &
  \textbf{$macro F1$} &
  Emotion Acc. &
  \textbf{$F1_{pos}$} &
  \textbf{$F1_{neg}$} &
  \textbf{$macro F1$} \\ \hline 
\multirow{18}{*}{Fold1 (DD)} & \multirow{3}{*}{Fold1 (DD)}   & RoBERTa-base  & -     & 56.64 & 85.13 & 70.88 & -     & 64.28 & \textbf{88.74} & 76.51 \\ \cline{3-11} 
&   & RoBERTa-large & -     & 50.48 & \textbf{87.35} & 68.91 & -     & 66.23 & 87.89 & 77.06 \\ \cline{3-11} 
&  & \subtwo          & 83.24 & \textbf{59.18} & 84.20 & \textbf{71.69} & 84.90 & \textbf{69.20} & 85.90 & \textbf{77.55} 
\\ \cline{3-11} 
&  & \overall & 80.12 & 53.03 & 86.80 & 69.91 & 82.02 & 64.90 & 88.12 & 76.51
\\ \cline{2-11} 
& \multirow{3}{*}{Fold2 (DD)}   & RoBERTa-base  & -     & 57.50 & 82.71 & 70.11 & -     & 59.06 & 86.91 & 72.98 \\ \cline{3-11} 
 &  & RoBERTa-large & -     & 56.13 & \textbf{88.33} & \textbf{72.23} & -     & 60.09 & \textbf{88.00} & \textbf{74.04} \\ \cline{3-11} 
&  & \subtwo          & 80.20 & \textbf{60.78} & 82.96 & 71.87 & 76.20 & \textbf{64.12} & 81.31 & 72.71 
\\ \cline{3-11} 
&  & \overall  & 75.56 & 55.23 & 86.12 & 70.67 & 72.87 & 58.43 & 87.21 & 72.82

\\ \cline{2-11} 
 & \multirow{3}{*}{Fold3 (DD)}   & RoBERTa-base  & -     & 57.52 & 82.72 & 70.12 & -     & 49.30 & 79.27 & 64.29 \\ \cline{3-11} 
& & RoBERTa-large & -     & 56.04 & \textbf{88.28} & \textbf{72.16} & -     & \textbf{60.63} & \textbf{88.30} & \textbf{74.46} \\ \cline{3-11} 
& & \subtwo          & 82.40 & \textbf{59.06} & 82.10 & 70.58 & 84.70 & 49.74 & 56.50 & 53.12 
\\ \cline{3-11} 
&  & \overall          & 81.34 & 56.77 & 86.29 & 71.53 & 80.19 & 46.43 & 88.13 & 67.28

\\ \cline{2-11} 
 & \multirow{3}{*}{Fold1 (IEMO)} & RoBERTa-base  & -     & 25.98 & 90.73 & 58.36 & -     & 28.02 & 95.67 & 61.85 \\ \cline{3-11} 
  &  & RoBERTa-large & -     & \textbf{32.34} & \textbf{95.61} & \textbf{63.97} & -     & \textbf{40.83} & \textbf{95.68} & \textbf{68.26} \\ \cline{3-11} 
&  & \subtwo          & 26.12 & 26.40 & 91.50 & 58.95 & 18.02 & 39.64 & 92.51 & 66.07 
\\ \cline{3-11} 
&  & \overall          & 25.34 & 25.23 & 89.52 & 57.37 & 17.92 & 36.54 & 92.84 & 64.69

\\ \cline{2-11} 
& \multirow{3}{*}{Fold2 (IEMO)} & RoBERTa-base  & -     & 32.60 & 89.99 & 61.30 & -     & 27.14 & 94.16 & 60.65 \\ \cline{3-11} 
  & & RoBERTa-large & -     & \textbf{36.61} & \textbf{94.60} & \textbf{65.60} & -     & 37.59 & \textbf{94.63} & 66.11 \\ \cline{3-11} 
 &  & \subtwo   & 30.21 & 32.20 & 90.52 & 61.36 & 15.87 & \textbf{42.41} & 92.40 & \textbf{67.40} 
 \\ \cline{3-11} 
&  & \overall          & 28.31 & 30.56 & 89.27 & 59.91 & 15.20 & 29.63 & 93.41 & 61.52
\\ \cline{2-11} 
    & \multirow{3}{*}{Fold3 (IEMO)} & RoBERTa-base  & -     & 33.24 & 90.30 & 61.77 & -     & 23.83 & 92.97 & 58.40 \\ \cline{3-11}  &                               & RoBERTa-large & -     & \textbf{36.55} & \textbf{94.59} & \textbf{65.57} & -     & \textbf{37.87} & \textbf{94.69} & \textbf{66.28} \\ \cline{3-11} 
   &   & \subtwo & 31.50 & 33.54 & 90.26 & 61.90 & 17.89 & 32.56 & 86.40 & 59.48 
   \\ \cline{3-11} 
&  & \overall          & 30.78 & 31.10 & 88.06 & 59.58 & 17.02 & 30.63 & 91.47 & 61.05
   \\ \hline

\end{tabular}
}
\vspace{0.5mm}
\caption{Results for Causal Emotion Entailment. Results are provided on RECCON-DD and RECCON-IEMO where RECCON-IEMO is only used during inference.}
\label{tab:sub2}
\vspace{-5mm}
\end{table*}

\noindent\textit{Results:} The results are shown in Table \ref{tab:task1_twostep}. For w/o CC and w/ CC, the $EM_{pos}$ and F$1_{pos}$ scores for \modelname\ are significantly higher in most of the cases. 
In general, we also noted that the high percentage of negative samples in the dataset affected the positive sample scores as well. Also, there seems to be a tradeoff between positive sample scores and negative sample scores. In some cases, our models are not able to beat the baselines for negative sample score, though this is not the case when we are performing inference on Fold2, Fold3, and IEMO (w/ CC) when the model is trained on Fold1. For these, our model surpasses the baseline. A possible reason might be that since, in these cases, we have the context of non-cause utterance combined with the target utterance, the model is easily able to distinguish if it is a positive or a negative sample. The model gets confused for negative samples when the non-cause utterances comes from the same dialogue (Fold1, w/ CC setting). For w/o CC, it is difficult for the model to identify the negative samples resulting in a lower negative sample score. F1 score is calculated as the utterance level mean of positive and negative samples, thus resulting in lower values since the dataset is unbalanced towards negative samples. For w/o CC, since the positive samples are the same for all the folds, the $EM_{pos}$ and $F1_{pos}$ are the same across all the folds. Only negative samples are different in these folds. 
We get good emotion prediction accuracy for Dailydialog. However, inference scores on IEMOCAP are low. Possible reason might be that since the training dataset (Dailydialog) and the inference dataset (IEMOCAP) are quite different, the model is not able to generalize well. Also, since IEMOCAP has some extra emotions, we clubbed similar kinds of emotions together, like happy and excited, anger and frustrated. Even with lower emotion scores, model performs better than the baselines, showing that the model is able to give better cause span results for cross-dataset settings as well. Results for training on Fold2 and Fold3 are shown in App. Table \ref{app-tab:sub1}. Similar trends were seen in these folds as well where for positive samples we are able to get better scores and for negative samples the scores drops. %\AB{similar trends, there isn't much different to fold1}

% \textbf{Results:}
% The results for Causal Emotion Entailment are shown in Table \ref{tab:sub2}. For the majority of the daily dialog dataset, our model was able to beat the baselines. But for IEMOCAP, the results are lower than the baselines, showing that for the task of causal emotion entailment, the model is not generalizing well. Here also, the emotion accuracies are for cross dataset (IEMOCAP) are lower. 
% With the inclusion of context, the F1 score increases showing the importance of context. Attention weights of the transformer-based encoder were also visualized, and it showed that since there are a lot of negative samples, the attention scores of the last layer were mostly corresponding to the first token that represents the negative sample (with start and end equal to zero) for the sample example. Figure shown in Appendix \ref{fig:attention_model_view}. Results for other folds are present in \ref{app:sub1}.

%-----------------------------------
%	SUBSECTION 2
%-----------------------------------

%\noindent\textbf{Task2: Causal Emotion Entailment:}

%%% ECPE results %%%%
% Please add the following required packages to your document preamble:
% \usepackage{multirow}

\begin{table}[t]\small
\centering
\renewcommand{\arraystretch}{1.5}
\setlength\tabcolsep{40pt}
\resizebox{\columnwidth}{!}{
\begin{tabular}{|c|c|c|c|c|}
\hline
\textbf{Dataset}      & \textbf{Model}            & \textbf{$F1_{pos}$} &
  \textbf{$F1_{neg}$} &
  \textbf{$macro F1$} \\ \hline
\multirow{7}{*}{DD}   & ECPE-2D                   & 55.50           & 94.96           & 75.23             \\ \cline{2-5}  & ECPE-MLL                  & 48.48           & 94.68           & 71.58             \\ \cline{2-5} & Rank CP                   & 33.00           & \textbf{97.30}           & 65.15             \\ \cline{2-5} 
 & RoBERTa-base              & 64.28           & 88.74           & 76.51             \\ \cline{2-5} 
 & RoBERTa-large             & 66.23           & 87.89           & 77.06             \\ \cline{2-5} 
 & \subtwo    & \textbf{69.20}           & 85.90           & \textbf{77.55}             \\ \cline{2-5} 
 & \overall & 64.90               & 88.12               & 76.51                 \\ \hline
\multirow{7}{*}{IEMOCAP} & ECPE-2D                   & 28.67           & \textbf{97.39}           & 63.03             \\ \cline{2-5} 
& ECPE-MLL                  & 20.23           & 93.55           & 57.65             \\ \cline{2-5} 
& Rank CP                   & 15.12           & 92.24           & 54.75             \\ \cline{2-5} 
& RoBERTa-base              & 28.02           & 95.67           & 61.85             \\ \cline{2-5} 
& RoBERTa-large             & \textbf{40.83}           & 95.68           & \textbf{68.26}             \\ \cline{2-5} 
& \subtwo    & 39.64           & 92.51           & 66.07             \\ \cline{2-5} 
 & \overall &        36.54         &     92.84  &    64.69               \\ \hline
\end{tabular}
}
\vspace{0.3mm}
%\caption{Comparison of Cause Entailment with previous baselines for Fold1 (With CC). }
\caption{Model and baseline results on Fold 1 (With CC) of Cause Entailment. }
\vspace{-4mm}
\label{tab:ecpe_subtwo}
\end{table}
%%%%%%%%%%%% Ablation subtask 2 %%%%%%

\begin{table}[t]\small
\centering
\renewcommand{\arraystretch}{1.5}
\setlength\tabcolsep{30pt}
\resizebox{\columnwidth}{!}{
\begin{tabular}{|c|c|cc|cc|}
\hline
\multirow{2}{*}{\textbf{Dataset}} & \multirow{2}{*}{\textbf{Model}} & \multicolumn{2}{c|}{\textbf{w/o CC}}        & \multicolumn{2}{c|}{\textbf{w/ CC}}         \\ \cline{3-6} 
 &        & \multicolumn{1}{c|}{\subone}   & \subtwo   & \multicolumn{1}{c|}{\subone}   & \subtwo   \\ \hline
\multirow{2}{*}{DD}      & w/ EP                  & \multicolumn{1}{c|}{62.86} & 58.36 & \multicolumn{1}{c|}{65.96} & 68.44  \\ \cline{2-6} 
 & w/o EP & \multicolumn{1}{c|}{59.21} & 52.68 & \multicolumn{1}{c|}{61.21} & 64.89  \\ \hline
\multirow{2}{*}{IEMOCAP} & w/ EP                  & \multicolumn{1}{c|}{51.42} & 25.77 & \multicolumn{1}{c|}{52.94} & 38.10  \\ \cline{2-6} 
 & w/o EP & \multicolumn{1}{c|}{50.21} & 24.49 & \multicolumn{1}{c|}{51.33} & 32.26 \\ \hline
\end{tabular}
}
\vspace{0.5mm}
\caption{\textbf{Ablation study.} w/ EP: with emotion prediction, w/o EP: without emotion prediction. The results are shown for $F1_{pos}$ (\%) for both the tasks.}
\label{tab:ablation}
\vspace{-5mm}
\end{table}

\noindent\textbf{Causal Emotion Entailment:}
\citet{poria2020recognizing} solve this task as a natural language inference task using RoBERTa-base and RoBERTa-large with linear layer on top.
% Evaluation metrics used are $F1_{pos}$, $F1_{neg}$ and $macro F1$. % (details in Appendix \ref{app:metrics}).

%%%%% Balanced dataset (task1) %%%%%%%%
% Please add the following required packages to your document preamble:
% \usepackage{multirow}
\begin{table*}[h]
\centering
\renewcommand{\arraystretch}{1.5}
\setlength\tabcolsep{5pt}
\resizebox{\columnwidth}{!}{

\begin{tabular}{|c|c|c|ccccc|ccccc|}
\hline
\multirow{2}{*}{\textbf{Train Fold}} &
  \multirow{2}{*}{\textbf{Test Fold}} &
  \multirow{2}{*}{\textbf{Model}} &
  \multicolumn{5}{c|}{\textbf{w/o CC}} &
  \multicolumn{5}{c|}{\textbf{w/ CC}} \\ \cline{4-13} 
 &
   &
   &
  \multicolumn{1}{c|}{Emotion Acc.} &
  \multicolumn{1}{c|}{\textbf{$EM_{pos}$}} &
  \multicolumn{1}{c|}{\textbf{$F1_{pos}$}} &
  \multicolumn{1}{c|}{\textbf{$F1_{neg}$}} &
  \textbf{F1} &
  \multicolumn{1}{c|}{Emotion Acc.} &
  \multicolumn{1}{c|}{\textbf{$EM_{pos}$}} &
  \multicolumn{1}{c|}{\textbf{$F1_{pos}$}} &
  \multicolumn{1}{c|}{\textbf{$F1_{neg}$}} &
  \textbf{F1} \\ \hline
\multirow{9}{*}{Fold1 (DD)} &
  \multirow{3}{*}{Fold1 (DD)} &
  RoBERTa-base &
  \multicolumn{1}{c|}{-} &
  \multicolumn{1}{c|}{36.54} &
  \multicolumn{1}{c|}{63.77} &
  \multicolumn{1}{c|}{\textbf{70.35}} &
   63.18 &
  \multicolumn{1}{c|}{-} &
  \multicolumn{1}{c|}{38.28} &
  \multicolumn{1}{c|}{68.8} &
  \multicolumn{1}{c|}{\textbf{83.48}} &
  \textbf{73.98}
   \\ \cline{3-13} 
 &
   &
  SpanBERT &
  \multicolumn{1}{c|}{-} &
  \multicolumn{1}{c|}{\textbf{36.96}} &
  \multicolumn{1}{c|}{64.84} &
  \multicolumn{1}{c|}{69.67} &
  \textbf{65.08}   &
  \multicolumn{1}{c|}{-} &
  \multicolumn{1}{c|}{38.70} &
  \multicolumn{1}{c|}{68.83} &
  \multicolumn{1}{c|}{81.54} &
  72.14
   \\ \cline{3-13} 
 &
   &
  \subone &
  \multicolumn{1}{c|}{80.23} &
  \multicolumn{1}{c|}{36.50} &
  \multicolumn{1}{c|}{\textbf{67.90}} &
  \multicolumn{1}{c|}{65.78} &
  60.13 &
  \multicolumn{1}{c|}{81.78} &
  \multicolumn{1}{c|}{\textbf{39.91}} &
  \multicolumn{1}{c|}{\textbf{72.41}} &
  \multicolumn{1}{c|}{72.61} &
  65.60 
  \\ \cline{3-13} 
 &
   &
  \overall &
  \multicolumn{1}{c|}{78.22} &
  \multicolumn{1}{c|}{35.08} &
  \multicolumn{1}{c|}{65.63} &
  \multicolumn{1}{c|}{67.89} &
  59.60 &
  \multicolumn{1}{c|}{79.43} &
  \multicolumn{1}{c|}{38.21} &
  \multicolumn{1}{c|}{70.56} &
  \multicolumn{1}{c|}{73.47} &
  64.13
   \\ \cline{2-13} 
 &
  \multirow{3}{*}{Fold2 (DD)} &
  RoBERTa-base &
  \multicolumn{1}{c|}{-} &
  \multicolumn{1}{c|}{36.54} &
  \multicolumn{1}{c|}{63.77} &
  \multicolumn{1}{c|}{57.39} &
  53.85 &
  \multicolumn{1}{c|}{-} &
  \multicolumn{1}{c|}{38.17} &
  \multicolumn{1}{c|}{68.56} &
  \multicolumn{1}{c|}{95.91} &
  85.21
   \\ \cline{3-13} 
 &
   &
  SpanBERT &
  \multicolumn{1}{c|}{-} &
  \multicolumn{1}{c|}{\textbf{36.96}} &
  \multicolumn{1}{c|}{64.84} &
  \multicolumn{1}{c|}{\textbf{66.80}} &
   \textbf{61.88} &
  \multicolumn{1}{c|}{-} &
  \multicolumn{1}{c|}{38.01} &
  \multicolumn{1}{c|}{68.98} &
  \multicolumn{1}{c|}{\textbf{96.24}} &
  \textbf{85.42}
   \\ \cline{3-13} 
 &
   &
  \subone &
  \multicolumn{1}{c|}{79.85} &
  \multicolumn{1}{c|}{36.50} &
  \multicolumn{1}{c|}{\textbf{67.90}} &
  \multicolumn{1}{c|}{66.14} &
  60.32 &
  \multicolumn{1}{c|}{66.43} &
  \multicolumn{1}{c|}{37.30} &
  \multicolumn{1}{c|}{\textbf{70.80}} &
  \multicolumn{1}{c|}{95.70} &
  84.29 
  \\ \cline{3-13} 
 &
   &
  \overall &
  \multicolumn{1}{c|}{75.45} &
  \multicolumn{1}{c|}{35.08} &
  \multicolumn{1}{c|}{65.63} &
  \multicolumn{1}{c|}{66.54} &
  60.23 &
  \multicolumn{1}{c|}{74.87} &
  \multicolumn{1}{c|}{\textbf{38.79}} &
  \multicolumn{1}{c|}{69.87} &
  \multicolumn{1}{c|}{95.41} &
  83.96
   \\ \cline{2-13} 
 &
  \multirow{3}{*}{Fold3 (DD)} &
  RoBERTa-base &
  \multicolumn{1}{c|}{-} &
  \multicolumn{1}{c|}{36.54} &
  \multicolumn{1}{c|}{63.77} &
  \multicolumn{1}{c|}{55.64} &
   52.83 &
  \multicolumn{1}{c|}{-} &
  \multicolumn{1}{c|}{38.17} &
  \multicolumn{1}{c|}{68.56} &
  \multicolumn{1}{c|}{95.85} &
  85.35
   \\ \cline{3-13} 
 &
   &
  SpanBERT &
  \multicolumn{1}{c|}{-} &
  \multicolumn{1}{c|}{\textbf{36.96}} &
  \multicolumn{1}{c|}{64.84} &
  \multicolumn{1}{c|}{58.93} &
   56.55 &
  \multicolumn{1}{c|}{-} &
  \multicolumn{1}{c|}{38.01} &
  \multicolumn{1}{c|}{68.98} &
  \multicolumn{1}{c|}{95.99} &
  85.15
   \\ \cline{3-13} 
 &
   &
  \subone &
  \multicolumn{1}{c|}{81.40} &
  \multicolumn{1}{c|}{36.50} &
  \multicolumn{1}{c|}{\textbf{67.90}} &
  \multicolumn{1}{c|}{64.74} &
   \textbf{60.56} &
  \multicolumn{1}{c|}{80.74} &
  \multicolumn{1}{c|}{37.30} &
  \multicolumn{1}{c|}{\textbf{70.80}} &
  \multicolumn{1}{c|}{\textbf{96.41}} &
  \textbf{85.70}
  \\ \cline{3-13} 
 &
   &
  \overall &
  \multicolumn{1}{c|}{80.04} &
  \multicolumn{1}{c|}{35.08} &
  \multicolumn{1}{c|}{65.63} &
  \multicolumn{1}{c|}{\textbf{65.32}} &
  58.98 &
  \multicolumn{1}{c|}{81.97} &
  \multicolumn{1}{c|}{\textbf{38.79}} &
  \multicolumn{1}{c|}{69.87} &
  \multicolumn{1}{c|}{96.03} &
  84.78
   \\ \hline

\end{tabular}
}
%\vspace{0.5mm}
\caption{Results for Cause Span Extraction task for the balanced dataset.} 
\label{tab:sub1_balanced}
\end{table*}

\noindent\textit{Results:} The results for Causal Emotion Entailment are shown in Table \ref{tab:sub2}. For the majority of the Dailydialog dataset, our models surpass the baselines. But for IEMOCAP, the results are lower than the baselines, showing that for the task of causal emotion entailment, the model is not generalizing well. RoBERTa-large gives significantly higher scores for IEMOCAP dataset showing that large-pretrained models work well in cross-dataset setting. %Results for training on Fold2 and Fold3 are shown in Appendix Table \ref{app-tab:sub2}. 
The results for training Fold2 and Fold3 (App. Table \ref{app-tab:sub2}) are consistent to Fold1 where RoBERTa-large shows significant improvements on IEMOCAP dataset. Attention weights of the transformer-based encoder were also visualized (App. Fig. \ref{app-fig:attention_model_view}), and it showed that since there are a lot of negative samples, the attention scores of the last layer were mostly corresponding to the first token that represents the negative sample (with start and end equal to zero) for the sample example. Table \ref{tab:ecpe_subtwo} shows the comparison with other baselines for Cause Entailment task for Fold 1 (with CC).
\section{Analysis} \label{sec:analysis}
%\subsection{Ablation Study}
\vspace{-3mm}
\noindent\textbf{Ablation Study:} To understand the importance of the auxiliary task of emotion prediction, we tried training the model with and without emotion prediction. Table \ref{tab:ablation} shows a performance drop when we don't train the model on the auxiliary task of emotion prediction. The study is performed on Fold1 (train and test). Ablation results across all the metrics are shown in App. \ref{app:ablation}. Since the results across training in Fold2 and Fold3 are similar to Fold1, we perform the ablation using Fold1 only. For IEMO it can be seen that the difference isn't significant. That might be because the emotion prediction in itself isn't much accurate due to different emotion label distribution of both RECCON-DD and RECCON-IEMO (details in App. Table \ref{app-tab:fold1_emotion_data}).
To understand the effect of \textit{beam size} (\S \ref{sec:mutec_infer}) on the SQuAD $F1_{pos}$ score for Cause Span Extraction, we experimented with different \textit{beam size}. After beam size of 3 (refer Fig. \ref{fig:beam} in Appendix), $F1_{pos}$ remains almost constant, thus we considered the beam size of 3 for our experiments. %Fig. \ref{fig:beam} shows the effect of \textit{beam size} on $F1_{pos}$ for Fold1 (w/o CC).

\noindent\textbf{Experiments with Balanced Dataset:}
The RECCON dataset contains lot more negative samples (data statistics in App. \ref{app:reccon}). We conducted same set of experiments with balanced dataset by reducing the number of negative samples, considering only two non-cause utterances for creating negative samples for each utterance (the statistics of balanced dataset in App. Table \ref{app-tab:balanced}). The results for balanced dataset are presented in Table \ref{tab:sub1_balanced} for CSE task. Comparing the results of before and after balancing the dataset, it is evident that reducing negative samples increases the overall score of positive samples for both the tasks. Thus having a balanced set of samples helps the model to learn better. The results of balanced dataset on CEE task is shown in Table \ref{tab:sub2_balanced} in Appendix. Our model showed similar trends as in the full dataset and gave good performance for positive samples and produced comparatively lower scores for negative samples for training and testing on similar folds. % setting. 

%The original dataset contained a lot of negative samples, as seen from the dataset statistics in Appendix \ref{app:reccon}. We ran the same set of experiments with a balanced dataset by reducing the number of negative samples considering only two non-cause utterances for creating negative samples for each utterance. The statistics of balanced dataset is shown in Appendix \ref{app:reccon} The results for balanced dataset are presented in Table \ref{tab:sub1_balanced} and Table \ref{tab:sub2_balanced} respectively.

% Comparing the results of before and after balancing the dataset, it is evident that reducing negative samples increased the overall score of positive samples for both the tasks. Thus having a balanced set of samples will help the model to learn better. Our model showed similar trends as in the full dataset and gave good performance for positive samples and produced comparatively lower scores for negative samples for training and testing on similar folds setting.

\section{Conclusion and Future Work}
\vspace{-3mm}
In this paper, we explore the task of extracting emotion cause in conversations. We experiment with the RECCON dataset. %, the first utterance level emotion causes annotation dataset. 
We propose a set of model architectures that do not require emotion annotations at the inference time. In particular we propose, multi-task learning approach where emotions are learned as an auxiliary task during cause span extraction (CSE) or causal emotion entailment (CEE) tasks. We also propose an overall end-to-end architecture for learning both the tasks together. As shown in experiments, the models give  comparable to better results without explicit emotion annotations at inference time. For future work, including the causal reasoning along with the cause spans in the annotated dataset can help the model to understand why this particular cause was selected. Also, currently, the RECCON dataset only uses dyadic conversations. This motivates the creation of datasets that use the multi-party setting. % as well. 

%The tasks of Causal Span Extraction and Causal Emotion Entailment in conversations are discussed and explored. A new multi-task learning approach where emotions are learned during training is discussed. We perform a comparison with the baselines for all the different dataset settings and are able to produce comparable to better results without explicit emotion annotations at inference time. For future work, including the causal reasoning alongside the cause spans in the annotated dataset can help the model to understand why this particular cause was selected. Also, currently, the RECCON dataset only uses dyadic conversations. There is scope to annotate datasets that use the multi-party setting as well.

\section{Acknowledgements}
We would like to thank reviewers for their insightful comments. This research is supported by SERB India (Science and Engineering Board) Research Grant number SRG/2021/000768. 

\bibliographystyle{plainnat}
\bibliography{anthology}

\clearpage
\newpage
\appendix
\section*{Appendix}

\section{RECCON Statistics}\label{app:reccon}

\textbf{RECCON} dataset was build using two popular conversational datasets \textbf{DialyDialog} \cite{li-etal-2017-dailydialog} and \textbf{IEMOCAP} \cite{busso2008iemocap}, both already had utterance level emotions associated. The RECCON dataset used only a subset of the IEMOCAP dataset and randomly selected dialogues from the DailyDialog dataset containing a minimum of four \textit{non-neutral} utterances. They did it because about 83\% of the DailyDialog dataset has \texttt{Neutral} labels. The annotated dataset was named RECCON-IE and RECCON-DD for IEMOCAP and DailyDialog, respectively.

% --------------------- utterance length -------------

\begin{figure*}[h]%
\centering
\subfigure{%
\label{app-fig:utlen-dd}%
\includegraphics[width=0.47\linewidth]{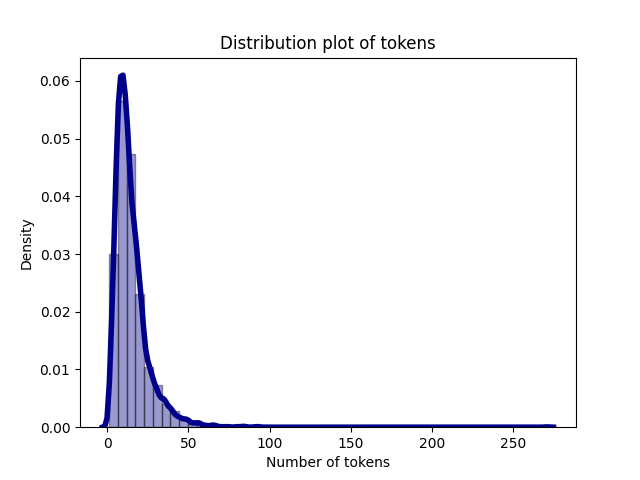}}%
\qquad
\subfigure{%
\label{app-fig:utlen-ie}%
\includegraphics[width=0.47\linewidth]{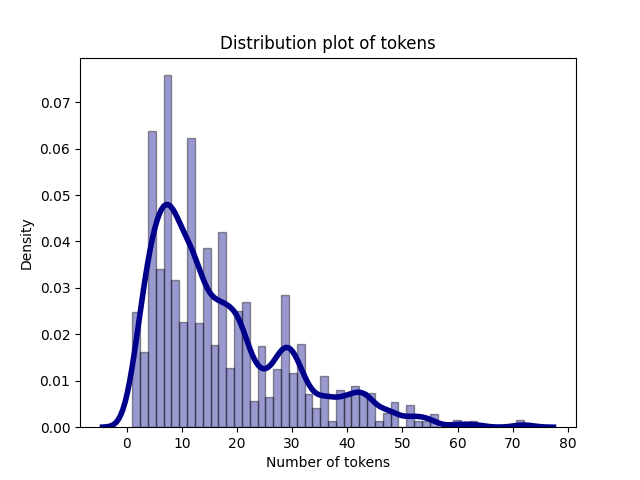}}%
\caption{\textbf{Distribution plot for number of tokens in an utterance of RECCON-DD and RECCON-IEMO}}
\end{figure*}

Table \ref{app-tab:datas} shows some of the statistics of RECCON annotated dataset. From the table, it can be seen that in RECCON-IE 40.5\% of utterances have a cause of the emotion in greater than three utterance distance in the conversational history, whereas in RECCON-DD only 13\% of utterances have their emotion cause in greater than three distance in the conversational history.  Fig. \ref{app-fig:utlen-dd} and \ref{app-fig:utlen-ie} shows the distribution of utterance length in RECCON-DD and RECCON-IEMO respectively.

% --------------------- statistics of reccon -------------

\begin{table}[!h]\small
\centering
\renewcommand{\arraystretch}{1.5}
\setlength\tabcolsep{30pt}
\resizebox{\columnwidth}{!}{
\begin{tabular}{||c|c|c||}
\hline
\textbf{\begin{tabular}[c]{@{}c@{}}Description\\ (Number of)\end{tabular}}                                                    & \textbf{RECCON-DD} & \textbf{RECCON-IE} \\ \hline \hline
Dialogues  & 1106 & 16 \\ \hline
Utterances  & 11104              & 665                \\ \hline
Utterances annotated with emotion cause  & 5861               & 494                \\ \hline
Utterances cater to background cause   & 395                & 70                 \\ \hline
\begin{tabular}[c]{@{}c@{}}Utterances where cause solely lies in the \\ same utterance\end{tabular}    & 1521               & 80                 \\ \hline
\begin{tabular}[c]{@{}c@{}}Utterance where cause includes the same \\ utterance along with contextual utterances\end{tabular} & 3370               & 243                \\ \hline \hline
\textbf{Number of emotion Utterances} & \textbf{}          & \textbf{}          \\ \hline
Anger & 451                & 89                 \\ \hline
Fear & 74                 & -                  \\ \hline
Disgust & 140                & -                  \\ \hline
Frustrated & -                  & 109                \\ \hline
Happiness & 4361               & 58                 \\ \hline
Sadness & 351                & 70                 \\ \hline
Surprise & 484                & -                  \\ \hline
Excited & -                  & 197                \\ \hline
Neutral & 5243               & 142                \\ \hline \hline
$U_t$ having cause at $U_{(t-1)}$  & 2851               & 183                \\ \hline
$U_t$ having cause at $U_{(t-2)}$ & 1182               & 124                \\ \hline
$U_t$ having cause at $U_{(t-3)}$ & 578                & 94                 \\ \hline
$U_t$ having cause at  \textgreater{} $U_{(t-3)}$ & 769                & 200                \\ \hline
\end{tabular}
}
\vspace{1mm}
\caption{\textbf{Statistics of RECCON annotated dataset}. Taken from the paper\cite{poria2020recognizing}}
\label{app-tab:datas}
\end{table}

Table \ref{app-tab:pos_neg_dataset} shows the final statistics of the dataset with positive and negative samples.

\begin{table}[!hbtp]\small
\centering
\renewcommand{\arraystretch}{1.5}
\setlength\tabcolsep{30pt}
\resizebox{\columnwidth}{!}{
\begin{tabular}{c c c c c c}
\multicolumn{3}{c}{\textbf{Data}} &
  \textbf{Train} &
  \textbf{Val} &
  \textbf{Test} \\ \hline \hline
\multirow{4}{*}{\textbf{\begin{tabular}[c]{@{}c@{}}Fold\\ 1\end{tabular}}} &
  \multirow{2}{*}{\textbf{DD}} &
  Positive Samples &
  7269 &
  347 &
  1894 \\ \cline{3-6} 
 &
   &
  Negative Samples &
  20646 &
  838 &
  5330 \\ \cline{2-6} 
 &
  \multirow{2}{*}{\textbf{IEMO}} &
  Positive Samples &
  - &
  - &
  1080 \\ \cline{3-6} 
 &
   &
  Negative Samples &
  - &
  - &
  11305 \\ \hline
\multirow{4}{*}{\textbf{\begin{tabular}[c]{@{}c@{}}Fold\\ 2\end{tabular}}} &
  \multirow{2}{*}{\textbf{DD}} &
  Positive Samples &
  7269 &
  347 &
  1184 \\ \cline{3-6} 
 &
   &
  Negative Samples &
  18428 &
  800 &
  4396 \\ \cline{2-6} 
 &
  \multirow{2}{*}{\textbf{IEMO}} &
  Positive Samples &
  - &
  - &
  1080 \\ \cline{3-6} 
 &
   &
  Negative Samples &
  - &
  - &
  7410 \\ \hline
\multirow{4}{*}{\textbf{\begin{tabular}[c]{@{}c@{}}Fold\\ 3\end{tabular}}} &
  \multirow{2}{*}{\textbf{DD}} &
  Positive Samples &
  7269 &
  347 &
  1894 \\ \cline{3-6} 
 &
   &
  Negative Samples &
  18428 &
  800 &
  4396 \\ \cline{2-6} 
 &
  \multirow{2}{*}{\textbf{IEMO}} &
  Positive Samples &
  - &
  - &
  1080 \\ \cline{3-6} 
 &
   &
  Negative Samples &
  - &
  - &
  7410 \\ 
\end{tabular}
}
\vspace{1mm}
\caption{\textbf{Dataset Statistics with both Positive Samples and Negative Samples.} DD refers to RECCON-DD and IEMO refers to RECCON-IEMO. Latent emotion cause is ignored.}
\label{app-tab:pos_neg_dataset}
\end{table}

\begin{table}[!hbtp]\small
\centering
\renewcommand{\arraystretch}{1.5}
\setlength\tabcolsep{30pt}
\resizebox{\columnwidth}{!}{
\begin{tabular}{|ccc|c|c|c|}
\hline
\multicolumn{3}{|c|}{\textbf{Data}} &
  \textbf{Train} &
  \textbf{Val} &
  \textbf{Test} \\ \hline
\multicolumn{1}{|c|}{\multirow{2}{*}{\textbf{Fold 1}}} &
  \multicolumn{1}{c|}{\multirow{2}{*}{\textbf{DD}}} &
  Positive Samples &
  7269 &
  347 &
  1894 \\ \cline{3-6} 
\multicolumn{1}{|c|}{} &
  \multicolumn{1}{c|}{} &
  Negative Samples &
  7356 &
  308 &
  1811 \\ \hline
\multicolumn{1}{|c|}{\multirow{2}{*}{\textbf{Fold2}}} &
  \multicolumn{1}{c|}{\multirow{2}{*}{\textbf{DD}}} &
  Positive Samples &
  7269 &
  347 &
  1184 \\ \cline{3-6} 
\multicolumn{1}{|c|}{} &
  \multicolumn{1}{c|}{} &
  Negative Samples &
  9124 &
  400 &
  2198 \\ \hline
\multicolumn{1}{|c|}{\multirow{2}{*}{\textbf{Fold 3}}} &
  \multicolumn{1}{c|}{\multirow{2}{*}{\textbf{DD}}} &
  Positive Samples &
  7269 &
  347 &
  1894 \\ \cline{3-6} 
\multicolumn{1}{|c|}{} &
  \multicolumn{1}{c|}{} &
  Negative Samples &
  9124 &
  400 &
  2198 \\ \hline
\end{tabular}
}
\vspace{1mm}
\caption{\textbf{Dataset Statistics for balanced set.} Negative samples were reduced by only taking two of the non-cause utterances for each target utterance instead of all the non-cause utterances for creating negative samples.}
\label{app-tab:balanced}
\end{table}

{\label{app:data_emotion}
\begin{table*}[t]\small
\centering
\renewcommand{\arraystretch}{1.3}
\setlength\tabcolsep{10pt}
\resizebox{\columnwidth}{!}{
\begin{tabular}{|c|c|c|c|c|c|c|c|c|c|}
\hline
\multicolumn{2}{|c|}{\textbf{Dataset}} &
  \textbf{Happiness} &
  \textbf{Surprise} &
  \textbf{Anger} &
  \textbf{Sadness} &
  \textbf{Disgust} &
  \textbf{Fear} &
  \textbf{Excited} &
  \textbf{Frustrated} \\ \hline
\multirow{3}{*}{DD} & Train & 22095 & 2205 & 1513 & 1269 & 555 & 278 & -   & -    \\ \cline{2-10} 
& Valid & 785   & 112  & 139  & 114  & 10  & 25  & -    & -    \\ \cline{2-10} 
& Test  & 4520  & 576  & 982  & 806  & 192 & 148 & -    & -    \\ \hline
IEMO                & Test  & 1295  & -    & 1535 & 1503 & -   & -   & 5778 & 2274 \\ \hline
\end{tabular}
}
\vspace{1mm}
\caption{\textbf{Number of emotion labels in Fold1 after the dataset is transformed into $U_t, U_i$ pairs.} Dataset is highly unbalanced and the distribution of emotion labels in DD and IEMO are not the same.}
\label{app-tab:fold1_emotion_data}
\end{table*}
}

\section{RECCON Tasks}
\label{app:reccon_tasks}
The task for \textbf{Cause Span Extraction} was solved as Question Answering task.
\textbf{Context:} Conversational history is context of a target utterance $U_t$. $(U_t, U_i)$ is used for a negative examples, where $U_i \notin C(U_t)$, conversational history of $U_t$. \\
\textbf{Question:} The following is how the question is phrased: \textit{“The target utterance is $<U_t>$. The evidence utterance is $<U_i>$. What is the causal span from evidence in the context that is relevant to the target utterance’s emotion $<E_t>$?}”\\
\textbf{Answer:} The causal span present in $U_i$ if $U_i \in C(U_t)$. $S$ is assigned an empty string for negative samples.\\
For \textbf{Fold2} and \textbf{Fold3} the context of $U_i$ is considered for negative samples.

For \textbf{Causal Emotion Entailment}, the task was solved as Natural language inference task. All the folds were structured as the 
follows:\\
Input as: \textit{$<E_t> <SEP> <U_t> <SEP> <U_i> <SEP> <H(U_t)>$}
and a label 1 if $U_i \in C(U_t)$ and label 0 if $U_i \notin C(U_t)$

For all folds in \textbf{Without Conversational Context (w/o CC)} setting, the context was not considered in the dataset.

\noindent We converted the data into the following:\\
\textbf{Task 1:} [id, emotion, $U_t$, $U_i$, cause\_span, context]\\
\textbf{Task 2:} [id, emotion, $U_t$, $U_i$, context, labels]

After transforming the dataset in $U_t, U_i$ pairs, Table \ref{app-tab:fold1_emotion_data} shows the number of emotion labels associated with Fold1.

\section{Model Hyperparameters} \label{app:hyperparams}

The value of hyperparameters used for both the tasks are listed in Table \ref{app-tab:hyperparams}

\begin{table}[!h]\small
\centering
\renewcommand{\arraystretch}{1.5}
\setlength\tabcolsep{30pt}
\resizebox{\columnwidth}{!}{
\begin{tabular}{|c|c|c|}
\hline 
\textbf{Hyperparameters} & \textbf{Task 1} & \textbf{Task 2} \\ \hline \hline
Number of Epochs         & \multicolumn{2}{c|}{12}           \\ \hline
Batch size               & \multicolumn{2}{c|}{16}           \\ \hline
\begin{tabular}[c]{@{}c@{}}Max. sequence length\\ (with context)\end{tabular} &
  \multicolumn{2}{c|}{512} \\ \hline
\begin{tabular}[c]{@{}c@{}}Max. sequence length\\ (without context)\end{tabular} &
  \multicolumn{2}{c|}{200} \\ \hline
Initial Learning Rate    & \multicolumn{2}{c|}{4e-5}         \\ \hline
Optimizer                & \multicolumn{2}{c|}{AdamW}        \\ \hline
Schedular &
  \multicolumn{2}{c|}{\begin{tabular}[c]{@{}c@{}}get\_linear\_schedule\_with\_warmup\\ (warmup steps = 4)\end{tabular}} \\ \hline
Max. answer length       & 200             & -               \\ \hline
n\_hidden\_states        & 12              & 4               \\ \hline
Dropout &
  \begin{tabular}[c]{@{}c@{}}Multi-sample dropout\\ (probability=0.5, layers=5)\end{tabular} &
  \begin{tabular}[c]{@{}c@{}}Simple dropout\\ (probability=0.1)\end{tabular} \\ \hline
Beam\_width              &  3              & -               \\ \hline
Bi-LSTM hidden dim       & -               & 384             \\ \hline
weight decay &   \multicolumn{2}{c|}{0.001}            \\ \hline

%   &                &              \\ \hline

%   &               &              \\ \hline

\end{tabular}
}
\vspace{1mm}
\caption{\textbf{Hyperparameter values used for both the tasks.}}
\label{app-tab:hyperparams}
\end{table}

%%%%%%%%%%%%%%%%%%%%%%%%

\section{Evaluation Metrics}\label{app:metrics}
\subsection{Cause Span Extraction Metrics}
For the evaluation of the models, the following metrics are used:

\noindent\textbf{Exact Match ($EM_{pos}$):} Exact Match corresponds to the percentage of exactly matched predicted spans to the gold spans.\\
\textbf{Positive F1 ($F1_{pos}$)}: SQuAD F1 score \cite{rajpurkar2016squad} calculated over positive examples. This metric measures the average overlap between the predicted spans and the ground span. 
\begin{gather}
    P_{pos} = \frac{Number\ of\ same\ tokens}{Number\ of\ predicted\ token}\\
    R_{pos} = \frac{Number\ of\ same\ tokens}{Number\ of\ gold\ tokens}\\
    F1_{pos} = \frac{2*P_{pos}*R_{pos}}{(P_{pos} + R_{pos})}
\end{gather}
\textbf{Negative F1 ($F1_{neg}$)}: F1 score calculated over negative examples. Here, the gold spans are empty spans.
\begin{gather}
    P_{neg} = \frac{Number\ of\ same\ empty\ spans}{Total\ number\ of\ predicted\ empty\ spans}\\
    R_{neg} = \frac{Number\ of\ same\ empty\ spans}{Total\ number\ of\ gold\ empty\ spans}\\
    F1_{neg} = \frac{2*P_{neg}*R_{neg}}{(P_{neg} + R_{neg})}
\end{gather}
\textbf{F1}: Overall F1 is calculated for each of the examples (positive and negative), which is followed by averaging over both of them.

\subsection{Causal Emotion Entailment Metrics}
\textbf{Positive F1 ($F1_{pos}$)}: F1 Score calculated for positive examples i.e., F1 score when positive samples are considered as true class.\\
\textbf{Negative F1 ($F1_{neg}$)}: F1 Score calculated for negative examples i.e., F1 score when negative samplesa are considered as true class.\\
\textbf{Macro F1:} Mean of class-wise (positive and negative) F1-scores.

{\label{app:sub1}
% ----------- Task 1 (fold2 and fold3) ------------------ %
\begin{table*}[!t]
\renewcommand{\arraystretch}{1.5}
\setlength\tabcolsep{5pt}
\resizebox{\columnwidth}{!}{
\begin{tabular}{|c|c|c|c|c|c|c|c|c|c|c|c|c|}
\hline
\multirow{2}{*}{\textbf{Train Fold}} &
  \multirow{2}{*}{\textbf{Test Fold}} &
  \multirow{2}{*}{\textbf{Model}} &
  \multicolumn{5}{c|}{\textbf{Without cc}} &
  \multicolumn{5}{c|}{\textbf{With cc}} \\ \cline{4-13} 
 &
   &
   &
  \textbf{Emotion Acc.} &
  \textbf{$EM_{pos}$} &
  \textbf{$F1_{pos}$} &
  \textbf{$F1_{neg}$} &
  \textbf{$F1$} &
  \textbf{Emotion Acc.} &
  \textbf{$EM_{pos}$} &
  \textbf{$F1_{pos}$} &
  \textbf{$F1_{neg}$} &
  \textbf{$F1$} \\ \hline \hline
%% Fold 2 %%
\multirow{24}{*}{Fold2(DD)} & \multirow{4}{*}{Fold1 (DD)}   & RoBERTa-base  & 
- & 33.26 & 58.44 & 71.29 & 60.45 & - & 36.06 & 65.04 & 0.19 & 17.12 \\ 
\cline{3-13} 
&    & SpanBERT & 
- & 32.31 & 58.61 & 72.52 & 61.70 & - & 31.52 & 60.81 & 0.67 & 16.19      \\ \cline{3-13} 
&  & Two Step &      
76.78 & 31.57  & 55.61  & \textbf{79.63}      & \textbf{68.12}  & 76.13 & 35.80 & \textbf{66.38}   & 0.75  & \textbf{17.68}   \\ 
\cline{3-13} 
&  & \subone &      
79.72 & \textbf{35.06} & \textbf{64.10} & 60.86 & 50.87 & 78.19 & 32.15 & 61.31   & \textbf{2.19}  & 16.89   \\ 

\cline{2-13} 
& \multirow{4}{*}{Fold2 (DD)}  
& RoBERTa-base  & 
- & 33.26 & 58.44 & 90.14 & 82.19 & - & 41.61 & 73.57 & 99.98 & 92.04 \\ \cline{3-13} 
&  & SpanBERT &
- & 32.31 & 58.61 & 90.20 & 82.29 & - & 41.97 & 74.85 & 99.94 & 92.43 \\ \cline{3-13} 
&  & Two Step  &  
79.95 & \textbf{35.37} &  63.13 &  86.17 & 75.81 & 80.09 & 41.29 & 74.31 & 99.95 & 92.23  \\
\cline{3-13} 
&  & \subone  &  
79.77 & 35.06 &  \textbf{64.10} &  81.91 & 71.05 & 78.30 & \textbf{42.56} & 74.62 & 99.91 & 92.31  \\
\cline{2-13} 
& \multirow{4}{*}{Fold3 (DD)} 
& RoBERTa-base  & 
- & -  & -     & -     & -     & -  &  -     & -     & -     & -     \\ \cline{3-13} 
&  & SpanBERT & 
- &     -  & -     & -     & -     & - &   -    & -     & -     & -     \\ \cline{3-13} 
&  & Two Step &     
79.95 & 35.37  & 63.13  &  83.13 &  72.59 &  80.09  & 41.29 & 74.31 & 99.79 & 92.01       \\ 
\cline{3-13} 
&  & \subone &     
79.85 & 35.06  & 64.10 & 79.23 &  68.25 &  79.15  & 42.56 & 74.62 & 99.75 & 92.09       \\
\cline{2-13} 
& \multirow{4}{*}{Fold1 (IEMO)} 
& RoBERTa-base  &
- & 15.93 & 31.74 & 90.70 & 82.91 & - & 22.96 & 46.87 & 4.66 & 6.35 \\ \cline{3-13} 
&  & SpanBERT &
- & 22.13 & 38.84 & 85.03 & 74.34 & - & 21.85 & 49.18 & 6.36 & 7.40 \\ \cline{3-13} 
& & Two Step  &     
22.43 & 22.69  & 40.35 & 84.01 & 72.69 & 22.55 & 28.43 & 50.30 & \textbf{43.96} & \textbf{30.72} \\
\cline{3-13} 
& & \subone  &     
21.43 & \textbf{30.28}  &  \textbf{50.68} & 72.90 & 58.64 &   20.10 & \textbf{30.28} & \textbf{58.19} & 6.36 & 8.09 \\
\cline{2-13} 
& \multirow{3}{*}{Fold2 (IEMO)} 
& RoBERTa-base  & 
- & 15.93 & 31.74 & 92.93 & 86.50 & - & 30.28 & 59.14 & 99.43 & 94.58      \\ \cline{3-13} 
&  & SpanBERT & 
- &  22.13 & 38.84 & 90.37 & 82.49 & - & 32.50 & 65.45 & 98.37 & 95.50       \\ 
\cline{3-13} 
&  & Two Step &   
23.67 & 28.98 & \textbf{51.5}  & 82.72   & 75.01  & 19.75 & 34.44 &  58.55 &  97.60 & 93.70       \\ 
\cline{3-13} 
&  & \subone &   
22.02 & \textbf{30.28} & 50.68  & 80.61  & 68.58  & 20.69 & \textbf{43.52} &  \textbf{77.71} &  98.01 & 94.21       \\
\cline{2-13} 
& \multirow{4}{*}{Fold3 (IEMO)} 
& RoBERTa-base  & 
- &    -   & -     & -     & -     & - &  -     & -     & -     & -     \\ \cline{3-13} 
& & SpanBERT &
- &    -   & -     & -     & -     & - & -      & -     & -     & -     \\ \cline{3-13} 
& & Two Step    &
23.67 & 28.98 & 51.5 & 81.91 & 74.7 & 20.77 & 34.44  & 58.55 & 97.11 & 92.87 \\ 
\cline{3-13} 
& & \subone    &
22.29 & 30.28 & 50.68 & 81.26 & 69.42 & 19.55 & 43.52  & 77.71 & 96.63 & 91.91 \\ 
\hline
%% fold 3 %%
\multirow{24}{*}{Fold3(DD)} & \multirow{3}{*}{Fold1 (DD)}   & RoBERTa-base  & 
- & 28.72 & 51.32 & 75.55 & 64.31 & - & 37.22 & 69.64 & 0.90 & 18.59 \\ 
\cline{3-13} 
&    & SpanBERT & 
- & 30.62 & 54.96 & 75.49 & 64.46 & - & 31.94 & 60.81 & 0.15 & 16.00      \\ 
\cline{3-13} 
&  & Two Step &      
78.11 &  32.37 & 59.15 & 67.5  & 56.1 &  76.13 &  31.36 & 61.63 & 0.71 & 16.35  \\ 
\cline{3-13} 
&  & \subone &      
80.75 &  \textbf{37.43} & \textbf{66.21} & 53.7  & 45.76 &  84.44 &  34.16 & 64.29 & \textbf{2.41} & 17.75  \\ 
\cline{2-13} 
& \multirow{3}{*}{Fold2 (DD)}  
& RoBERTa-base  & 
- & -  & -     & -     & -     & -  &  -     & -     & -     & -     \\ \cline{3-13} 
&  & SpanBERT & 
- &     -  & -     & -     & -     & - &   -    & -     & -     & -   \\ \cline{3-13} 
&  & Two Step  &  
79.44 & 32.37 &  58.95 &    87.36   & 77.24  &  79.09   & 40.34 & 74.55 & 99.93   &     92.27  \\
\cline{3-13} 
&  & \subone  &  
79.49 & 37.43 &  66.21 & 76.24 & 65.53  &  71.26  & 41.24 & 74.31 & 99.90  &  92.23 \\
\cline{2-13} 
& \multirow{3}{*}{Fold3 (DD)} 
& RoBERTa-base  & 
- & 28.72 & 51.32 & 90.06 & 82.11 & - & 41.29 & 74.95 & 99.94 & 92.44     \\ \cline{3-13} 
&  & SpanBERT & 
- &  30.62 & 54.96 & 89.41 & 81.21 & - & 42.61 & 75.36 & 99.93 & 92.46    \\ \cline{3-13} 
&  & Two Step &     
79.44 & 32.37 & 58.95   & 86.66  & 76.34 & 81.12  &  40.34 &  74.55 & 99.85    &  92.16     \\ 
\cline{3-13} 
&  & \subone &     
80.49 & \textbf{37.43} & \textbf{66.21} & 74.62 & 63.99 & 92.52  &  41.02 &  75.08 & 99.80    &  92.29     \\
\cline{2-13} 
& \multirow{4}{*}{Fold1 (IEMO)} 
& RoBERTa-base  &
- & 14.54 & 26.51 & 92.33 & 85.61 & - & 21.20 & 48.34 & 11.42 & 9.76 \\ \cline{3-13} 
&  & SpanBERT &
- & 17.41 & 31.75 & 89.41 & 80.94 & - & 21.48 & 45.49 & 4.01 & 5.84 \\ \cline{3-13} 
& & Two Step  &     
23.44 & 26.39  & 44.38 & 82.67  & 70.95 & 22.55 & 26.30  & 44.9  &   \textbf{42.16} & \textbf{28.9}         \\ 
\cline{3-13} 
& & \subone  &     
22.40 & \textbf{36.30}  & \textbf{57.54} & 70.00  & 55.61 & 21.21 & \textbf{32.59}  & \textbf{59.47} & 6.46  &  8.24 \\
\cline{2-13} 
& \multirow{4}{*}{Fold2 (IEMO)} 
& RoBERTa-base  & 
- &    -   & -     & -     & -     & - &  -     & -     & -     & -     \\ \cline{3-13} 
& & SpanBERT &
- &    -   & -     & -     & -     & - & -      & -     & -     & -     \\ \cline{3-13} 
&  & Two Step &   
24.64 & 26.3  & 44.71   & 88.69   & 79.76  & 23.12  & 36.48     &  59.05 &  97.36 & 93.41   \\ 
\cline{3-13} 
&  & \subone &   
21.01 & 36.3 & 57.74  & 79.38  & 67.32  & 20.18  & 46.20     &  76.64 &  98.31 & 94.47   \\
\cline{2-13} 
& \multirow{3}{*}{Fold3 (IEMO)} 
& RoBERTa-base  & 
- & 14.54 & 26.51 & 93.68 & 87.79 & - & 24.35 & 53.46 & 97.84 & 94.08     \\ \cline{3-13} 
& & SpanBERT &
- &  17.41 & 31.75 & 91.85 & 84.86 & - & 32.87 & 62.70 & 99.54 & 95.11  
\\ \cline{3-13} 
& & Two Step    &
24.64 & 26.30 & 44.71 & 88.27 & 79.84 & 27.12 & 36.48 & 59.05 & 97.64 & 93.61 \\ 
 \cline{3-13} 
& & \subone    &
22.11 & \textbf{36.30} & \textbf{57.74} & 79.48 & 67.45 & 22.66 & \textbf{46.20} & \textbf{76.64} & 97.08 & 92.42 \\
 \hline

\end{tabular}
}
\vspace{1mm}
\caption{\textbf{Comparision results for Cause Span Extraction task} for Two Step and \subone model architecture on RECCON-DD and RECCON-IEMO. IEMO dataset is only used in the inference phase. (Fold2 and Fold3)}
\label{app-tab:sub1}
\end{table*}
}

\begin{figure*}[t]%
\centering
\subfigure{%
\label{app-fig:ut_start}%
\includegraphics[width=\columnwidth]{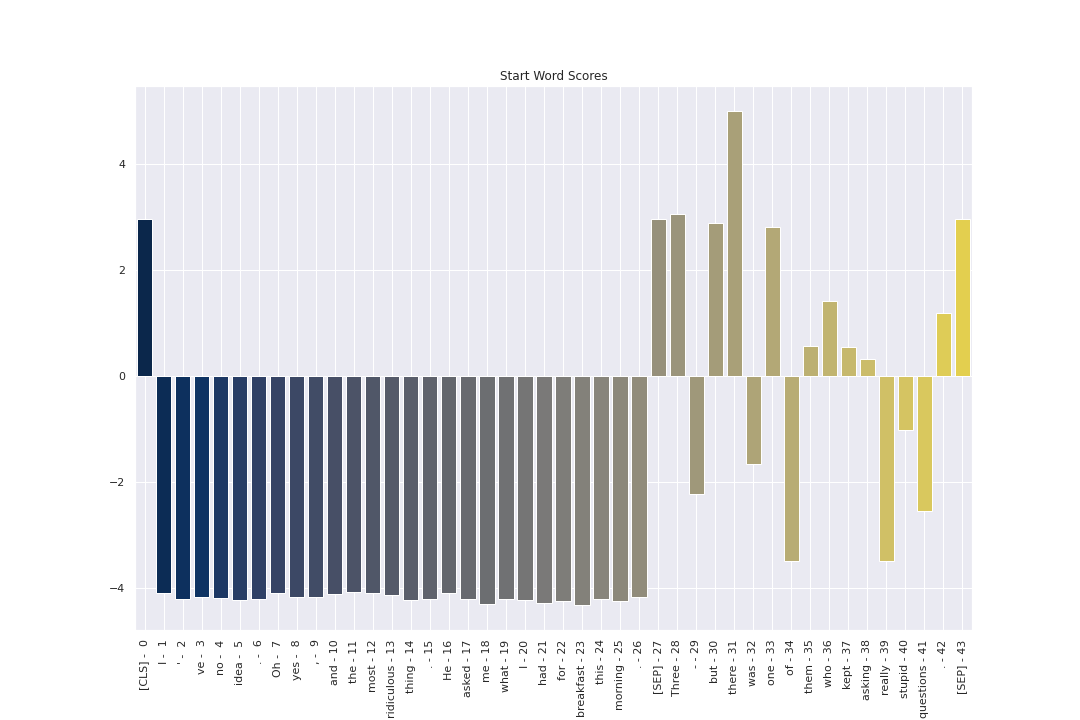}}%
\qquad
\subfigure{%
\label{app-fig:ut_end}%
\includegraphics[width=\columnwidth]{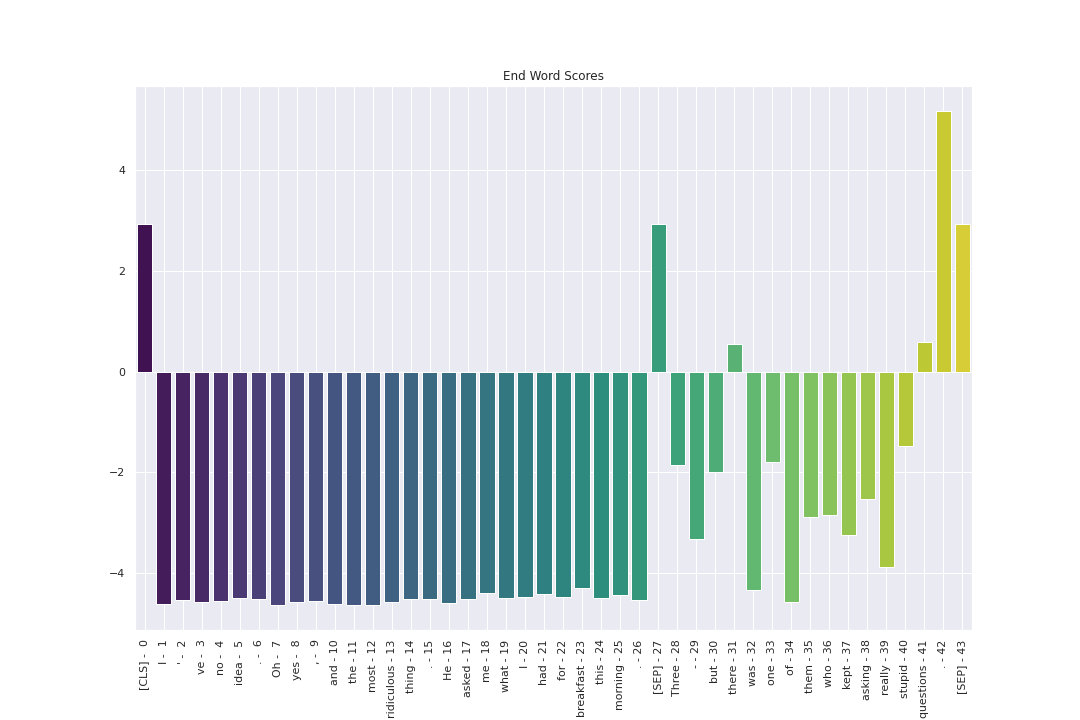}}%
\caption{\textbf{Start and End Words score for an example input}. \textit{there} is predicted as the start token and \textit{.} as the end token.}
\label{app-fig:ut_scores}
\end{figure*}

% \section{Appendices}
The logit scores (for start and end) calculated by the model for a test sample are given in Fig. \ref{app-fig:ut_scores}.

% --------------------- Logit scores for start and end tokens -------------

{\label{app:sub2}
\begin{table*}[t]\small
\renewcommand{\arraystretch}{1.10}
\setlength\tabcolsep{10pt}
\resizebox{\columnwidth}{!}{
\begin{tabular}{|c|c|c|c|c|c|c|c|c|c|c|}
\hline
\multirow{2}{*}{\textbf{Train Fold}} &
  \multirow{2}{*}{\textbf{Test Fold}} &
  \multirow{2}{*}{\textbf{Model}} &
  \multicolumn{4}{c|}{\textbf{Without cc}} &
  \multicolumn{4}{c|}{\textbf{With cc}} \\ \cline{4-11} 
 &
   &
   &
  \textbf{Emotion Acc.} &
  \textbf{$F1_{pos}$} &
  \textbf{$F1_{neg}$} &
  \textbf{$macro F1$} &
  \textbf{Emotion Acc.} &
  \textbf{$F1_{pos}$} &
  \textbf{$F1_{neg}$} &
  \textbf{$macro F1$} \\ \hline \hline

\multirow{18}{*}{Fold2(DD)}  & \multirow{3}{*}{Fold1 (DD)}   & RoBERTa-base  & -     & 52.52 & 75.51 & 64.02 & -     & 41.86 & 3.25  & 22.55 \\ \cline{3-11} 
                             &                               & RoBERTa-large & -     & 51.57 & 67.58 & 59.57 & -     & 43.25 & 19.95 & 31.60 \\ \cline{3-11} 
                             &                               & \subtwo          & 72.23 & \textbf{53.17} & \textbf{78.23} & \textbf{65.70} & 82.01 & 42.99 & 17.90 & 30.45 \\ \cline{2-11} 
                             & \multirow{3}{*}{Fold2 (DD)}   & RoBERTa-base  & -     & 76.21 & 91.23 & 83.72 & -     & 89.37 & 95.21 & 92.32 \\ \cline{3-11} 
                             &                               & RoBERTa-large & -     & 79.52 & 91.27 & 85.40 & -     & 93.05 & 97.22 & 95.13 \\ \cline{3-11} 
                             &                               & \subtwo          & 74.14 & 74.53 & 89.91 & 82.22 & 81.55 & \textbf{94.01} & \textbf{97.59} & \textbf{95.80} \\ \cline{2-11} 
                             & \multirow{3}{*}{Fold3 (DD)}   & RoBERTa-base  & -     & -     & -     & -     & -     & -     & -     & -     \\ \cline{3-11} 
                             &                               & RoBERTa-large & -     & -     & -     & -     & -     & -     & -     & -     \\ \cline{3-11} 
                             &                               & \subtwo          & 71.43 & 72.52 & 89.98 & 81.25 & 81.52 & 69.47 & 81.03 & 75.25 \\ \cline{2-11} 
                             & \multirow{3}{*}{Fold1 (IEMO)} & RoBERTa-base  & -     & 31.51 & 92.09 & 61.80 & -     & 25.22 & 74.69 & 49.96 \\ \cline{3-11} 
                             &                               & RoBERTa-large & -     & 29.64 & 87.68 & 58.66 & -     & 26.30 & 76.44 & 51.37 \\ \cline{3-11} 
                             &                               & \subtwo          & 18.5  & 30.74 & 90.55 & 60.65 & 15.97 & \textbf{27.44} & \textbf{80.21} & \textbf{53.82} \\ \cline{2-11} 
                             & \multirow{3}{*}{Fold2 (IEMO)} & RoBERTa-base  & -     & 46.12 & 93.80 & 69.96 & -     & 65.09 & 95.60 & 80.35 \\ \cline{3-11} 
                             &                               & RoBERTa-large & -     & 48.36 & 92.06 & 70.21 & -     & 61.12 & 95.59 & 78.35 \\ \cline{3-11} 
                             &                               & \subtwo          & 20.85 & \textbf{49.02} & 92.80 & \textbf{70.91} & 17.73 & 61.68 & 95.44 & 78.56 \\ \cline{2-11} 
                             & \multirow{3}{*}{Fold3 (IEMO)} & RoBERTa-base  & -     & -     & -     & -     & -     & -     & -     & -     \\ \cline{3-11} 
                             &                               & RoBERTa-large & -     & -     & -     & -     & -     & -     & -     & -     \\ \cline{3-11} 
                             &                               & \subtwo          & 18.91 & 61.17 & 95.62 & 78.39 & 20.08 & 46.09 & 92.35 & 69.22 \\ \hline
\multirow{18}{*}{Fold 3(DD)} & \multirow{3}{*}{Fold1 (DD)}   & RoBERTa-base  & -     & 52.02 & 74.59 & 63.31 & -     & 41.64 & 2.99  & 22.31 \\ \cline{3-11} 
                             &                               & RoBERTa-large & -     & 51.53 & 65.76 & 58.65 & -     & 41.86 & 4.89  & 23.38 \\ \cline{3-11} 
                             &                               & \subtwo          & 79.98 & \textbf{53.86} & \textbf{77.43} & \textbf{65.64} & 90.33 & \textbf{42.43} & \textbf{12.33} & \textbf{27.38} \\ \cline{2-11} 
                             & \multirow{3}{*}{Fold2 (DD)}   & RoBERTa-base  & -     & -     & -     & -     & -     & -     & -     & -     \\ \cline{3-11} 
                             &                               & RoBERTa-large & -     & -     & -     & -     & -     & -     & -     & -     \\ \cline{3-11} 
                             &                               & \subtwo          & 80.28 & 74.88 & 90.49 & 82.69 & 79.03 & 52.73 & 46.50 & 49.62 \\ \cline{2-11} 
                             & \multirow{3}{*}{Fold3 (DD)}   & RoBERTa-base  & -     & 74.73 & 90.33 & 82.53 & -     & 92.64 & 96.99 & 94.81 \\ \cline{3-11} 
                             &                               & RoBERTa-large & -     & 75.79 & 88.43 & 82.11 & -     & 93.34 & 97.23 & 95.29 \\ \cline{3-11} 
                             &                               & \subtwo          & 80.80 & 74.11 & 90.05 & 82.08 & 95.62 & \textbf{96.36} & \textbf{98.48} & \textbf{97.42} \\ \cline{2-11} 
                             & \multirow{3}{*}{Fold1 (IEMO)} & RoBERTa-base  & -     & 34.74 & 91.46 & 63.10 & -     & 19.13 & 54.25 & 36.69 \\ \cline{3-11} 
                             &                               & RoBERTa-large & -     & 27.58 & 84.13 & 55.86 & -     & 18.33 & 48.01 & 33.17 \\ \cline{3-11} 
                             &                               & \subtwo          & 18.56 & 30.52 & 90.04 & 60.28 & 24.27 & \textbf{19.90} & \textbf{56.53} & \textbf{38.21} \\ \cline{2-11} 
                             & \multirow{3}{*}{Fold2 (IEMO)} & RoBERTa-base  & -     & -     & -     & -     & -     & -     & -     & -     \\ \cline{3-11} 
                             &                               & RoBERTa-large & -     & -     & -     & -     & -     & -     & -     & -     \\ \cline{3-11} 
                             &                               & \subtwo          & 21.41 & 47.76 & 93.08 & 70.42 & 26.86 & 31.16 & 71.29 & 51.73 \\ \cline{2-11} 
                             & \multirow{3}{*}{Fold3 (IEMO)} & RoBERTa-base  & -     & 51.23 & 93.70 & 72.46 & -     & 63.91 & 94.55 & 79.23 \\ \cline{3-11} 
                             &                               & RoBERTa-large & -     & 43.00 & 88.47 & 65.74 & -     & 59.03 & 92.21 & 75.62 \\ \cline{3-11} 
                             &                               & \subtwo          &  21.86     &  48.30     &     93.42  &  70.86     & 30.07 & 58.02 & 93.59 & 75.74 \\ \hline
 \end{tabular}
}
\vspace{1mm}
\caption{\textbf{Comparison results for Causal Emotion Entailment.} Results are provided on RECCON-DD and RECCON-IEMO where RECCON-IEMO is only used during inference.}
\label{app-tab:sub2}
\end{table*}
}

\begin{figure*}[!h]
\centering
  \includegraphics[width=\linewidth]{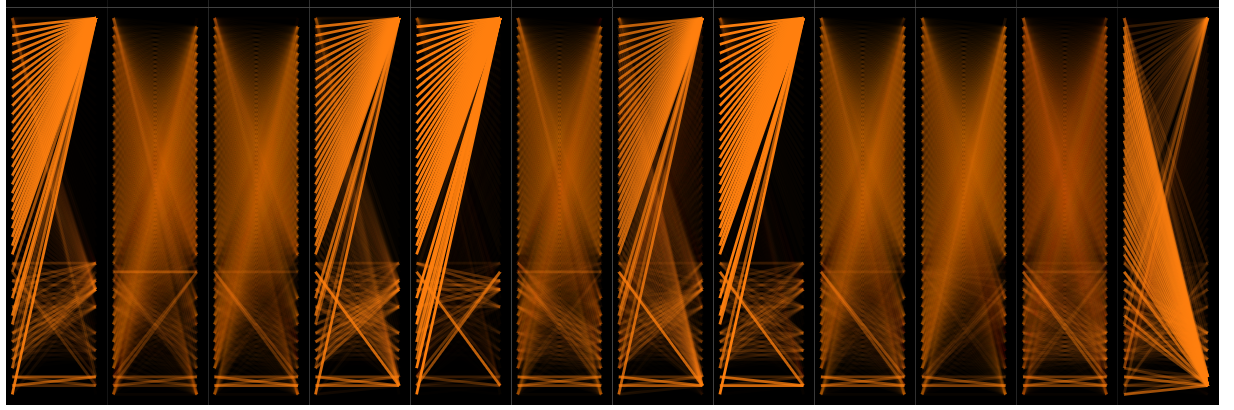}
  \caption{\textbf{Attention for Layer 12 of SpanBERT across all the attention heads for a trained model.} Bertviz \cite{vig-2019-multiscale} library was used to produce these attention visualizations on trained model (on Fold1) for task 1.}
  \label{app-fig:attention_model_view}
\end{figure*}

{\label{app:ablation}
\begin{table*}[t]\small
\renewcommand{\arraystretch}{1.5}
\setlength\tabcolsep{10pt}
\resizebox{\columnwidth}{!}{
\begin{tabular}{c c c c c c c c c c}
\hline
\multirow{2}{*}{\textbf{Dataset}} &
  \multirow{2}{*}{\textbf{Model Setting}} &
  \multicolumn{4}{c}{\textbf{Without cc}} &
  \multicolumn{4}{c}{\textbf{With cc}} \\ \cline{3-10} 
 &
  &
  \textbf{$EM_{pos}$} &
  \textbf{$F1_{pos}$} &
  \textbf{$F1_{neg}$} &
  $F1$ &
  \textbf{$EM_{pos}$} &
  \textbf{$F1_{pos}$} &
  \textbf{$F1_{neg}$} &
  $F1$ \\ \hline
\multirow{2}{*}{DD}   & With emotion prediction    & 36.06 & 62.86 & 62.50 & 52.10 & 36.43 & 65.96 & 75.07 & 63.64\\ \cline{2-10} 
                      & Without emotion prediction & 34.75 & 59.21 & 63.58 & 50.33 & 33.32 & 61.21 & 78.09 & 62.87\\ \hline
\multirow{2}{*}{IEMO} & With emotion prediction    & 31.85 & 51.42 & 69.20 & 54.42 & 30.56 & 52.94 & 82.24 & 70.37\\ \cline{2-10} 
                      & Without emotion prediction & 30.99 & 50.21 & 70.43 & 53.14 & 29.43 & 51.33 & 83.10 & 70.09 \\ \hline
\end{tabular}
}
\vspace{1mm}
\caption{Task 1 end-to-end model trained with and without the auxiliary emotion prediction task.}
%\caption{Task 1 end-to-end model trained on the auxiliary emotion prediction task vs. without trained on auxiliary emotion prediction task. Auxiliary emotion prediction task, in general, gives better scores, indicating the importance of this task. The study is performed on Fold1 (train and test). For IEMO it can be seen that the difference isn't significant; that might be because the emotion prediction in itself isn't much accurate due to different emotion label distribution of both RECCON-DD and RECCON-IEMO.}
\label{app-tab:sub1_ablation}
\end{table*}

% For Causal Emotion Entailment, an ablation study is conducted where the model is trained with and without the auxiliary task of emotion prediction. The results in Table \ref{tab:sub2_ablation} show that there is a performance drop when we don't train on the auxiliary task of emotion prediction.

\begin{table*}[t]\small
\renewcommand{\arraystretch}{1.5}
\setlength\tabcolsep{15pt}
\resizebox{\columnwidth}{!}{
\begin{tabular}{c c c c c c c c}
\multirow{2}{*}{\textbf{Dataset}} &
  \multirow{2}{*}{\textbf{Model Setting}} &
  \multicolumn{3}{c}{\textbf{Without cc}} &
  \multicolumn{3}{c}{\textbf{With cc}} \\ \cline{3-8} 
 &
   &
  \textbf{$F1_{pos}$} &
  \textbf{$F1_{neg}$} &
  macro $F1$ &
  \textbf{$F1_{pos}$} &
  \textbf{$F1_{neg}$} &
  macro $F1$ \\ \hline
\multirow{2}{*}{DD}   & With emotion prediction    & 58.26 & 84.94 & 71.60 & 68.44 & 86.88 & 77.66 \\ \cline{2-8} 
                      & Without emotion prediction & 52.68 & 79.69 & 66.19 & 64.89 & 84.37 & 74.63 \\ \hline
\multirow{2}{*}{IEMO} & With emotion prediction    & 25.77 & 90.39 & 58.08 & 38.10 & 93.50 & 65.80 \\ \cline{2-8} 
                      & Without emotion prediction & 24.49 & 88.00 & 56.25 & 32.26 & 92.32 & 62.29 
\end{tabular}
}
\vspace{1mm}
\caption{Task 2 end-to-end model trained with and without auxiliary emotion task.}
%\caption{Task 2 end-to-end model trained with and without auxiliary emotion task. Training on the auxiliary emotion task helps the model to learn good emotion representations, thus increasing the F1 score. The above study is performed on fold1 as both training and test set.}
\label{app-tab:sub2_ablation}
\end{table*}
}

\section{Ablation Study} \label{app:ablation}
We performed ablation study on Fold1 by removing the emotion predictor for both the tasks. The details are shown in Table \ref{app-tab:sub1_ablation} and \ref{app-tab:sub2_ablation} respectively.

% --------------------- Effect of beam width on Pos F1 score -------------
\begin{figure}[h]
\centering
  \includegraphics[scale=0.25]{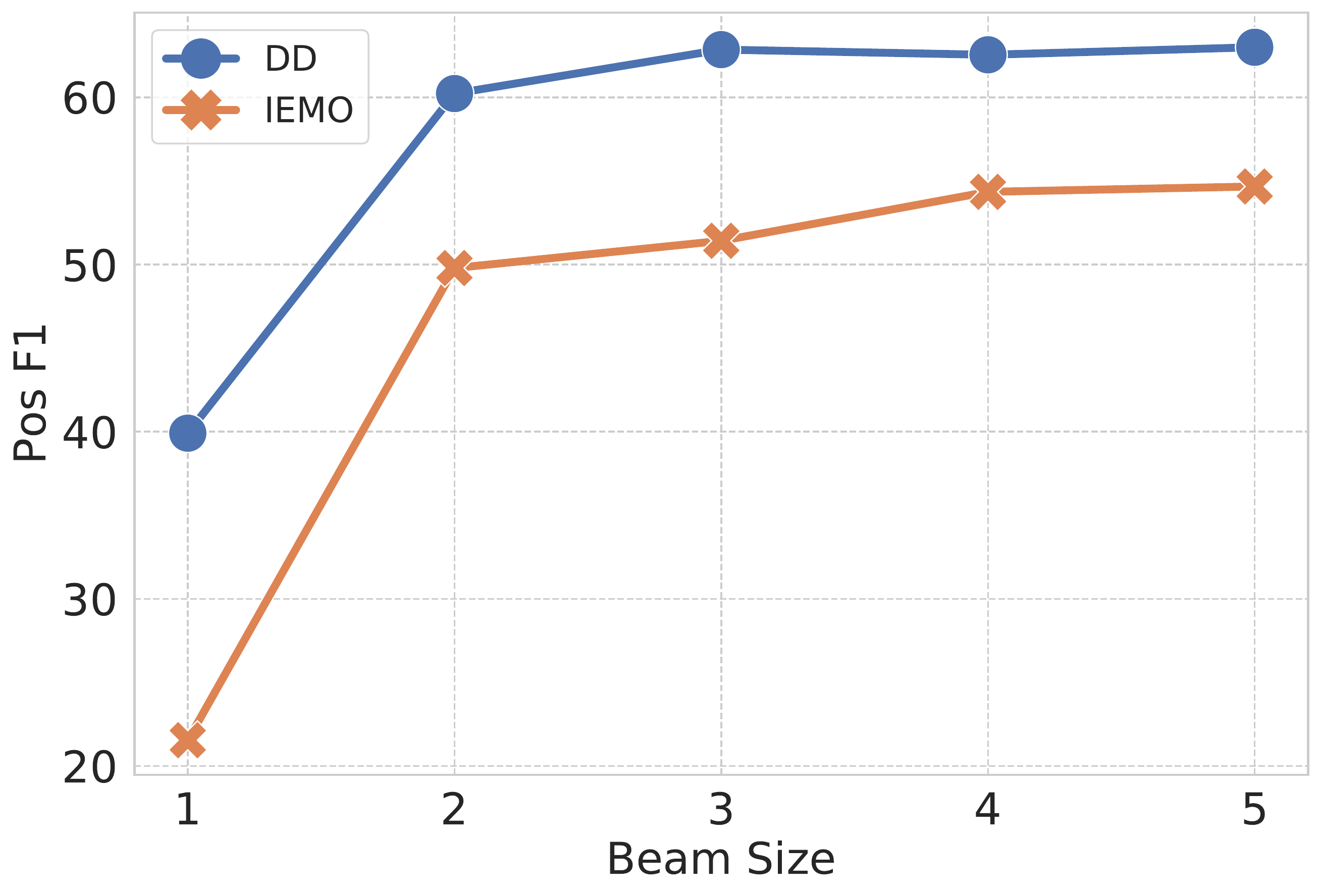}
  \caption{Effect of \textit{Beam Size} on $F1_{pos}$ (\%) for Fold1 (\textit{w/o CC}). }
  \label{fig:beam}
%   \vspace{-5mm}
\end{figure}

\begin{table*}[h]
\centering
\renewcommand{\arraystretch}{1.5}
\setlength\tabcolsep{8pt}
\resizebox{\columnwidth}{!}{
\begin{tabular}{|c|c|c|cccc|cccc|}
\hline
\multirow{2}{*}{\textbf{Train Fold}} &
  \multirow{2}{*}{\textbf{Test Fold}} &
  \multirow{2}{*}{\textbf{Model}} &
  \multicolumn{4}{c|}{\textbf{w/o CC}} &
  \multicolumn{4}{c|}{\textbf{w/ CC}} \\ \cline{4-11} 
 &
   &
   &
  \multicolumn{1}{c|}{Emotion Acc.} &
  \multicolumn{1}{c|}{\textbf{$F1_{pos}$}} &
  \multicolumn{1}{c|}{\textbf{$F1_{neg}$}} &
  macro F1 &
  \multicolumn{1}{c|}{Emotion Acc.} &
  \multicolumn{1}{c|}{\textbf{$F1_{pos}$}} &
  \multicolumn{1}{c|}{\textbf{$F1_{neg}$}} &
  macro F1 \\ \hline
\multirow{9}{*}{Fold1 (DD)} &
  \multirow{3}{*}{Fold1 (DD)} &
  RoBERTa-base &
  \multicolumn{1}{c|}{-} &
  \multicolumn{1}{c|}{75.67} &
  \multicolumn{1}{c|}{69.96} &
   72.82 &
  \multicolumn{1}{c|}{-} &
  \multicolumn{1}{c|}{85.12} &
  \multicolumn{1}{c|}{85.14} &
   85.13
   \\ \cline{3-11} 
 &
   &
  RoBERTa-large &
  \multicolumn{1}{c|}{-} &
  \multicolumn{1}{c|}{\textbf{75.95}} &
  \multicolumn{1}{c|}{69.73} &
  72.84 &
  \multicolumn{1}{c|}{-} &
  \multicolumn{1}{c|}{\textbf{85.43}} &
  \multicolumn{1}{c|}{84.93} &
  \textbf{85.18}
   \\ \cline{3-11} 
 &
   &
  \subtwo &
  \multicolumn{1}{c|}{79.14} &
  \multicolumn{1}{c|}{75.81} &
  \multicolumn{1}{c|}{\textbf{71.21}} &
  \textbf{73.51} &
  \multicolumn{1}{c|}{80.56} &
  \multicolumn{1}{c|}{85.13} &
  \multicolumn{1}{c|}{84.69} &
  84.90 
  \\ \cline{3-11} 
 &
   &
  \overall &
  \multicolumn{1}{c|}{78.22} &
  \multicolumn{1}{c|}{73.42} &
  \multicolumn{1}{c|}{70.16} &
  71.79 &
  \multicolumn{1}{c|}{79.43} &
  \multicolumn{1}{c|}{82.65} &
  \multicolumn{1}{c|}{\textbf{86.41}} &
  84.53
  \\ \cline{2-11} 
 &
  \multirow{3}{*}{Fold2 (DD)} &
  RoBERTa-base &
  \multicolumn{1}{c|}{-} &
  \multicolumn{1}{c|}{66.32} &
  \multicolumn{1}{c|}{55.22} &
   60.77 &
  \multicolumn{1}{c|}{-} &
  \multicolumn{1}{c|}{65.09} &
  \multicolumn{1}{c|}{\textbf{65.02}} &
  65.05
   \\ \cline{3-11} 
 &
   &
  RoBERTa-large &
  \multicolumn{1}{c|}{-} &
  \multicolumn{1}{c|}{67.79} &
  \multicolumn{1}{c|}{\textbf{58.12}} &
  \textbf{62.96} &
  \multicolumn{1}{c|}{-} &
  \multicolumn{1}{c|}{68.48} &
  \multicolumn{1}{c|}{54.20} &
  61.34
   \\ \cline{3-11} 
 &
   &
  \subtwo &
  \multicolumn{1}{c|}{75.85} &
  \multicolumn{1}{c|}{\textbf{69.55}} &
  \multicolumn{1}{c|}{53.04} &
  61.29 &
  \multicolumn{1}{c|}{75.77} &
  \multicolumn{1}{c|}{\textbf{69.84}} &
  \multicolumn{1}{c|}{54.98} &
  62.41 
  \\ \cline{3-11} 
 &
   &
  \overall &
  \multicolumn{1}{c|}{75.45} &
  \multicolumn{1}{c|}{65.21} &
  \multicolumn{1}{c|}{56.38} &
  60.79 &
  \multicolumn{1}{c|}{74.87} &
  \multicolumn{1}{c|}{68.41} &
  \multicolumn{1}{c|}{62.11} &
  \textbf{65.26}
  \\ \cline{2-11} 
 &
  \multirow{3}{*}{Fold3 (DD)} &
  RoBERTa-base &
  \multicolumn{1}{c|}{-} &
  \multicolumn{1}{c|}{66.13} &
  \multicolumn{1}{c|}{54.64} &
   60.39 &
  \multicolumn{1}{c|}{-} &
  \multicolumn{1}{c|}{57.14} &
  \multicolumn{1}{c|}{\textbf{43.15}} &
   \textbf{50.14}
   \\ \cline{3-11} 
 &
   &
  RoBERTa-large &
  \multicolumn{1}{c|}{-} &
  \multicolumn{1}{c|}{67.76} &
  \multicolumn{1}{c|}{58.04} &
   62.90 &
  \multicolumn{1}{c|}{-} &
  \multicolumn{1}{c|}{59.57} &
  \multicolumn{1}{c|}{13.71} &
  36.64 
   \\ \cline{3-11} 
 &
   &
  \subtwo &
  \multicolumn{1}{c|}{81.03} &
  \multicolumn{1}{c|}{\textbf{69.74}} &
  \multicolumn{1}{c|}{\textbf{59.44}} &
  \textbf{64.59} &
  \multicolumn{1}{c|}{83.41} &
  \multicolumn{1}{c|}{\textbf{60.54}} &
  \multicolumn{1}{c|}{23.41} &
  41.97 
  \\ \cline{3-11} 
 &
   &
  \overall &
  \multicolumn{1}{c|}{80.04} &
  \multicolumn{1}{c|}{66.28} &
  \multicolumn{1}{c|}{57.62} &
  61.95 &
  \multicolumn{1}{c|}{81.97} &
  \multicolumn{1}{c|}{56.23} &
  \multicolumn{1}{c|}{35.21} &
  45.72
  \\ \hline
\end{tabular}
}
%\vspace{0.5mm}
\caption{Results for Causal Emotion Entailment task for the balanced dataset.}
\label{tab:sub2_balanced}
\vspace{-5mm}
\end{table*}

\end{document}